\documentclass[a4paper]{cas-sc} 
\usepackage[numbers]{natbib}
\usepackage{caption}  
\usepackage{cleveref}
\usepackage{subfig}   
\usepackage{amsthm}
\usepackage[edges]{forest}
\usetikzlibrary{arrows.meta}
\tikzset{el/.style={font=\scriptsize, inner sep=1.5pt,
fill=white, rounded corners=1pt}}
\newcommand{\dzero}{d^{2}_{0}}
\newcommand{\dinf}{d^{2}_{\pm\infty}}

\usepackage{algorithm}
\usepackage{algpseudocode}

\usepackage{enumitem} 

\newcommand{\R}{\mathbb{R}}

\newcommand{\D}{\mathcal{D}}
\newcommand{\bdelta}{\boldsymbol{\delta}}
\newcommand{\bphi}{\boldsymbol\phi}
\newcommand{\bth}{\boldsymbol{\theta}}
\newcommand{\hth}{\boldsymbol{\hat{\theta}}}
\newcommand{\Fis}{\mathcal{I}}
\newcommand{\Fhat}{\widehat{\Fis}}
\newcommand{\Rcv}{\widehat{R}_{\mathrm{CV}}}
\newcommand{\scv}{\widehat{s}_{\mathrm{CV}}}
\newcommand{\dF}{d_c^{2}}
\newcommand{\KL}{D_{\mathrm{KL}}}
\DeclareMathOperator*{\argmin}{arg\,min}

\DeclareMathOperator{\sgn}{sgn}

\theoremstyle{definition}

\usepackage[normalem]{ulem}

\begin{document}

\let\WriteBookmarks\relax
\def\floatpagepagefraction{1}
\def\textpagefraction{.001}
\shorttitle{Fisher-Geometric Pruning for Symbolic Regression}
\shortauthors{V. Guidetti et~al.}

\title [mode = title]{From Overparameterised Expressions to Parsimonious Models: Fisher-Geometric Pruning for Symbolic Regression}                      
\author[1]{Veronica Guidetti}[orcid=0000-0003-2233-0097]
\cormark[1]
\ead{veronica.guidetti@unimore.it}

\author[2]{Luca La~Rocca}[orcid=0000-0003-0495-1308]
\ead{luca.larocca@unimore.it}

\author[3]{Fabr\'icio Olivetti~de~Fran\c{c}a}[orcid=0000-0002-2741-8736]
\ead{folivetti@ufabc.edu.br}

\author[4]{Gabriel Kronberger}[orcid=0000-0002-3012-3189]
\ead{gabriel.kronberger@fh-hagenberg.at}

\affiliation[1]{organization={Department of Physics, Informatics and Mathematics, University of Modena and Reggio Emilia},addressline={via G. Campi 213/b}, city={Modena}, postcode={41126},country={Italy}}

\affiliation[2]{organization={Department of Education and Humanities, University of Modena and Reggio Emilia}, addressline={viale Timavo 93}, city={Reggio nell'Emilia}, postcode={42121},country={Italy}}

\affiliation[3]{organization={Center for Mathematics, Computation and Cognition (CMCC), Federal University of ABC}, addressline={Avenida dos Estados, 5001}, city={Santo Andre}, postcode={09280-560}, country={Brazil}}

\affiliation[4]{organization={University of Applied Sciences Upper Austria, Digital Campus Hagenberg}, addressline={Softwarepark 11}, city={Hagenberg}, postcode={4232}, country={Austria}}

\cortext[cor1]{Corresponding author}

\begin{abstract}
Genetic programming symbolic regression (GPSR) often returns expressions that are larger than necessary, with dormant branches and finely tuned algebraic configurations that fit sampling noise.
Existing complexity-control strategies, such as multi-objective size-accuracy search or information criteria embedded in the fitness function, act on syntactic representations during model search.
We introduce Fisher-Geometric pruning (FGP), a post-hoc model-reduction framework that acts on fitted models and can complement these techniques.

FGP combines local information geometry at the maximum-likelihood solution with explicit compatibility tests for interpretable parametric boundaries.
A taxonomy of parametric directions separates local sensitivity from proximity to reducible boundaries, enabling the treatment of unidentifiable directions and reductions through zero constraints, asymptotic limits, and signed parameter identifications.
Candidate reductions are realised by symbolic contraction and refitting, then accepted only when their cross-validated predictive risk remains within a prescribed tolerance.

We evaluate FGP on controlled diagnostic problems and Pareto fronts generated by GPSR.
Across both settings, FGP substantially reduces structural complexity while preserving, and often improving, out-of-sample accuracy.
Comparisons with competing pruning strategies and ablations show that, by combining geometric screening, boundary compatibility, and predictive validation, FGP yields the most consistent joint gains in parsimony and generalisation.

\end{abstract}



\begin{keywords}
symbolic regression \sep  genetic programming \sep model pruning \sep Fisher information \sep information geometry \sep interpretability

\end{keywords}

\maketitle

\section{Introduction}
\label{sec:intro}
Symbolic regression (SR) searches a space of closed-form mathematical expressions for a model that explains an observed input-output relation \citep{koza1992gp,poli2008field,schmidt2009distilling}. Unlike black-box predictors, an SR model is an explicit formula, which makes SR attractive whenever a mechanistic reading of the fitted model matters, e.g., scientific discovery, engineering surrogates, regulated decision support. The dominant implementation of SR is based on genetic programming (GP): genetic programming symbolic regression (GPSR) evolves a population of expression trees by variation and selection. To improve the search through the parameterized functional space, modern GPSR systems couple combinatorial search of tree structures with nonlinear optimization of numerical coefficients
 \citep{topchy2001,kommenda2020parameter}.

The interpretability promised by SR is, however, conditional on the returned
expression being parsimonious enough to read, and GP is well known not to
deliver this for free. Evolved trees may grow without commensurate gains in
predictive accuracy, i.e., the classical bloat phenomenon
\citep{luke2006comparison,silva2009dynamic}, and, more insidiously, the final population may contain expressions whose apparent structure has little to do with the mechanism that generated the data.
Large fitted expressions may include redundant or sample-specific structure, such as branches effectively switched off by fitted coefficients, so that their apparent complexity substantially exceeds that of the effective model supported by the data.

Existing approaches to complexity control in GPSR follow two broad directions. The first acts during evolutionary search, for example through structural complexity objectives in multi-objective optimisation frameworks, parsimony pressure, or model-selection criteria \citep{vladislavleva2009order,defranca2023alleviating,Haut2025,poli2008}.
More refined approaches, based on minimum description length (MDL), explicitly account for both structural complexity and parameter precision \citep{kronberger2026guiding}, but still measure complexity through an explicit coding scheme.

A second direction acts post-hoc on fitted parametric models. Information-geometric model reduction, including the manifold boundary approximation method (MBAM) \citep{transtrum2014,transtrum2015} and Bayesian renormalisation (BR) \citep{berman2023,howard2025}, uses the geometry of the fitted model to identify reducible parameter combinations and construct lower-dimensional descriptions. More recently,
Fisher-geometric distances to zero-parameter boundaries have been proposed directly as a pruning criterion \citep{berman2026optimal}.

Both directions have limitations in the SR setting. Search-time complexity control evaluates candidate expressions in their current representation and may therefore fail to reveal a simpler effective model latent within a more complex evolved expression. Existing information-geometric approaches, by contrast, do not directly provide an automatic reduction procedure for heterogeneous, machine-generated symbolic expressions in which simplification may occur at qualitatively different parametric boundaries. They either rely on case-specific exploration of the model manifold, focus on weakly resolved directions, or restrict pruning to a particular boundary such as parameter zeroing.

As we show, recovering effective emergent models from evolved GPSR expressions requires recognising two facts: removable structure is not confined to the sloppy end of the Fisher spectrum, and parameter zeroing alone is not a sufficiently rich reduction mechanism for weakly identified directions. A locally sensitive direction may still be removable because the fitted solution lies close to a simplifying boundary, while a weakly sensitive direction may make a finite and stable contribution to the model. Likewise, flatness may arise from weak excitation, asymptotic saturation, or parameter symmetry, each calling for a different type of reduction. Thus, local sensitivity and proximity to a reducible boundary are distinct properties, and both are needed to determine which parts of a fitted symbolic model can be removed.

\paragraph{Contributions.} 
These observations motivate a post-hoc reduction framework that separates identifying potentially removable structure from realising and validating the corresponding simplification. Our contributions are threefold. 
\begin{enumerate}
\item \textbf{A taxonomy of parametric directions} for fitted symbolic models (\cref{sec:taxonomy}) built on two distinct local quantities: Fisher sensitivity and distance to an interpretable parametric boundary.
The taxonomy separates legitimate from removable structure at both the stiff and sloppy ends of the spectrum.

\item \textbf{Fisher-Geometric Pruning (FGP), a computationally lightweight reduction pipeline} (\cref{sec:method}) that turns the taxonomy into a concrete set of admissible symbolic moves that progressively prune fitted models.
Local geometry identifies parameters or parameter combinations whose structural role should be tested, while explicit boundary-compatibility tests determine whether the fitted model is consistent with a specific simplifying constraint or limit.
Candidate reductions are then refitted and accepted through a cross-validated predictive gate, using the Bayesian information criterion (BIC) only to break ties among predictively equivalent alternatives.

\item \textbf{A two-tier empirical evaluation}
(\cref{sec:setup,sec:results}) on ten controlled diagnostic problems with known reductions and on complete GPSR Pareto fronts across multiple noise levels and sample sizes. We compare FGP with a coefficient-magnitude pruning baseline and targeted ablations of candidate generation, reduction moves, and the acceptance rule. Across both settings, FGP provides the most consistent joint reduction in structural complexity and out-of-sample error.
The code and data needed to reproduce the experiments are available at \url{https://github.com/guidettiv/EmergentSR.git}.
\end{enumerate}

The paper is organized as follows.
\Cref{sec:related} reviews complexity control in SR and information-geometric model reduction, and positions our method with respect to both.
\Cref{sec:prelim} fixes notation and derives the local geometric objects used throughout.
\Cref{sec:taxonomy} introduces the taxonomy of parametric directions.
\Cref{sec:method} presents the FGP algorithm and its motivations.
\Cref{sec:setup} describes the experimental protocol, and \cref{sec:results} reports the results.
\Cref{sec:conclusion} discusses scope and limitations.
Detailed derivations, controlled-problem specifications, hyperparameter sensitivity analyses, and computational-cost comparisons are deferred to the appendices.
Additional experimental results and figures are provided in the supplementary material.
\section{Related work}
\label{sec:related}

\subsection{Complexity control in GP-based symbolic regression}
\label{sec:rel-gp}

Controlling expression growth is a fundamental concern of GP and GPSR, where unconstrained evolution can produce increasingly large trees without commensurate improvements in fitness \citep{koza1992gp,poli2008field}.
Operational responses fall into different groups. 

Generic GP approaches aim to control complexity at the level of the evolving representation.
Structural constraints limit tree depth or size, while parsimony pressure penalises large individuals during selection;
\citet{luke2006comparison} and \citet{silva2009dynamic} survey and compare
these mechanisms.

In GPSR, complexity control can also take advantage of the predictive nature of the task. 
Multi-objective approaches treat error and complexity as separate objectives and return a Pareto front of candidate models. Complexity has been measured by node count, by order of nonlinearity
\citep{vladislavleva2009order}, by recursive structural measures \citep{kommenda2015complexity}, and, in the transformation-interaction-rational family, jointly with an explicit overfitting-control objective
\citep{defranca2021interaction,defranca2023alleviating}. 

A related line of work treats complexity control as statistical model selection.
Akaike's information criterion (AIC) and BIC \citep{akaike1974new,schwarz1978estimating} penalize likelihood by the number of fitted parameters, while MDL-type criteria account for both structure and parameter precision
\citep{bartlett2024exhaustive,rissanen1978modeling,grunwald2007mdl}.
A recent comparative study of model-selection criteria for GPSR~\citep{kronberger2026guiding} highlights the importance of considering the function complexity and the required precision of the parameters. This distinction is particularly relevant for modern GPSR systems, which typically combine combinatorial search over expression structures with nonlinear least-squares optimization of their numerical parameters
\citep{kommenda2020parameter}.

These approaches control complexity during search or model selection, but assess candidate expressions in their current representation and parameterisation.
Post-hoc simplification offers a complementary perspective: algebraic methods, such as equality saturation with e-graphs \citep{defranca2025egraph}, reduce representational redundancy while preserving the represented function.
In contrast, our approach seeks data-supported reductions of fitted parametric models, potentially changing the represented function while preserving predictive performance.
The two forms of simplification are complementary and can be combined.

\subsection{Information-geometric model reduction}
\label{sec:rel-geo}

Information geometry treats a parametric statistical model as a Riemannian manifold
with the Fisher information as metric \citep{amari1997,amari2000methods,amari2016}.
For classical human-derived nonlinear least squares frameworks, the model manifold is typically ``sloppy'': the Fisher spectrum spans many orders of magnitude, the manifold is a hyper-ribbon with a hierarchy of widths, and only a few stiff combinations are resolved at a given experimental precision \citep{transtrum2011geometry,transtrum2015,Quinn2021}.
This structure explains why complex microscopic systems admit effective compact descriptions \citep{machta2013}.

MBAM \citep{transtrum2014,transtrum2015} exploits such a structure constructively
to reduce parametric model complexity: starting at the maximum-likelihood estimate (MLE),
it integrates a geodesic along the sloppiest eigen-direction until the model manifold reaches a boundary,
at which point one or more parameters attain a limiting value (typically $0$ or $\infty$)
and a reduced model is read off. 

Another approach is Bayesian renormalisation \citep{berman2023,howard2025}, which prunes parametric directions by inverting Bayesian inference: the Fisher spectrum at the MLE defines a hierarchy of inference scales, and a flow in model space integrates out the degrees of freedom that are unresolvable at a given scale, producing a coarse-grained effective model.
Given its highly non-trivial implementation in high-dimensional applications, which would require several marginalization steps, the flow is approximated by diagonal Fisher heuristics that shrink sloppy parameters towards zero.

More recently, information-geometric distances have been proposed directly for neural-network pruning, measuring the cost of parameter removal through the Fisher distance from the fitted model to the corresponding zero-parameter hypersurface \citep{berman2026optimal}.
This provides a geometric alternative to magnitude- and local-sensitivity-based pruning, but remains restricted to reductions that set a parameter to zero.

These approaches exploit Fisher geometry to identify reducible structure, but vary in the directions and boundaries they consider. FGP extends this perspective in two respects.
First, removable structure in machine-generated expressions need not correspond to the sloppiest direction: a locally influential component may still be compatible with a simplifying finite boundary.
Second, the relevant symbolic reduction need not be parameter zeroing; asymptotic limits and parameter identifications can remove structures that zero-based pruning cannot.
Both Bayesian renormalisation in its current empirical implementation and Fisher-distance pruning have been applied through zero-parameter removal, whereas our framework treats zero compatibility as only one possible reduction mechanism and handles non-identifiability separately.
MBAM, on the contrary, can explore both finite and asymptotic manifold boundaries, but its application is typically tailored to individual case studies and does not readily provide an automatic reduction procedure across heterogeneous model syntaxes.
It also requires repeated integration of geodesic equations, making direct application to large SR populations computationally demanding.
A comparable computational burden arises for a full Bayesian-renormalisation treatment, which would require repeated information-shell renormalisation steps rather than the diagonal heuristics used in practice.

\subsection{Parameter pruning, sparsity and identifiability}
\label{sec:rel-prune}

Removing parameters from a fitted model is a classical problem outside SR as well.
Magnitude pruning \citep{han2015learning,blalock2020state} zeroes the smallest coefficients and refits.
On top of being scale dependent, applying this method to nonlinear parameterisations may be misleading as a small coefficient need not indicate a negligible functional contribution.
Other approaches to model sparsification include penalty methods \citep{tibshirani1996lasso} and sparse dictionary methods for dynamical systems \citep{brunton2016sindy}.
However, these are most naturally formulated for a fixed set of candidate terms, often linear in their coefficient and therefore do not directly address the nested, nonlinear parametrisation generated by unrestricted SR expression trees.

Identifiability analysis supplies the vocabulary for the degeneracies we exploit in our taxonomy: structural non-identifiability, a redundancy of the parameterisation whose signature is a singular Fisher matrix, and empirical non-identifiability, a direction left unconstrained by finite noisy data \citep{raue2009structural}.
The machinery, however, does not transfer: many constructive structural-identifiability tools exploit polynomial or rational representations \citep{chis2011structural}, whereas GP expressions may freely compose exponentials, logarithms, and other transcendental functions.

Classical nested-model testing is likewise only partially suited: linear restrictions remain useful for removing unidentifiable or sloppy directions, but the deeply nested parameter dependence of symbolic trees means
the reductions we need are typically taken either at a singular Fisher matrix \citep{watanabe2009algebraic} or as asymptotic parameter limits, which are manifold boundaries rather than interior restrictions \citep{chernoff1954distribution,self1987asymptotic}.

\section{Preliminaries}
\label{sec:prelim}

We consider a regression problem with data
$\D=\{(x_j,y_j)\}_{j=1}^{n}$, $x_j\in\R^{d}$, $y_j\in\R$, and a parametric
model $f(\,\cdot\,;\bth):\R^{d}\to\R$ with $\bth\in\Theta\subseteq\R^{p}$.
Throughout, $f$ is the symbolic expression to be reduced and $\bth$ collects its
numerical parameters. We assume additive homoscedastic Gaussian noise:
\begin{equation}
y_j \;=\; f(x_j;\bth) + \varepsilon_j,
\qquad \varepsilon_j \sim\mathcal{N}(0,\sigma^{2}),
\qquad
\mbox{independently over $j=1,\ldots,n$},
\label{eq:noise}
\end{equation}
where $\sigma\ge0$ represents the noise level, so that the likelihood factorises as
\begin{equation}
\mathcal{L}(\bth,\sigma)
=\prod_{j=1}^{n}\frac{1}{\sqrt{2\pi\sigma^{2}}}
\exp\!\left\{-\frac{\big(y_j-f(x_j;\bth)\big)^{2}}{2\sigma^{2}}\right\},\quad \bth\in\Theta,\,\sigma\in\R_+.
\label{eq:lik}
\end{equation}
Maximising \cref{eq:lik} 
is equivalent to nonlinear least squares:
\begin{equation}
\hth = \argmin_{\bth\in\Theta}\sum_{j=1}^{n}\big(y_j-f(x_j;\bth)\big)^{2},
\qquad
\hat\sigma^{2}=\frac{1}{n}\sum_{j=1}^{n}\big(y_j-f(x_j;\hth)\big)^{2}.
\label{eq:mle}
\end{equation}
Throughout, $\hat\sigma$ denotes the residual standard deviation at the current
fit, not the true noise level $\sigma$ of \cref{eq:noise}; the two coincide only
asymptotically and only for a well-specified model.

\subsection{Fisher information and local geometry}
\label{sec:prelim-fisher}

The Fisher information matrix (FIM) is the natural local metric on a Riemannian model
manifold \citep{amari2000methods}. For the Gaussian model in \cref{eq:noise},
the expectation over $y_j$ can be taken in closed form and,
focusing on the parameter of interest $\bth$, the per-observation FIM reduces to
a Gauss--Newton form in the model Jacobian rather than in the score:
\begin{equation}
\Fis_j(\bth)
=\frac{1}{\sigma^{2}}\,
\nabla_{\bth}f(x_j;\bth)\,\nabla_{\bth}f(x_j;\bth)^{\!\top}.
\label{eq:fim-pop}
\end{equation}
For $n$ independent observations, the total information carried by the sample is
$\Fis_{[n]}(\bth)\;=\;\sum_{j=1}^n\Fis_j(\bth)$.
Its empirical counterpart at the MLE, which is the object actually computed,
is $\Fhat_{[n]}=\Fis_{[n]}(\hth)$, whose entries are given by
\begin{equation}
\Fhat_{[n]hk}
=\frac{1}{\hat\sigma^2}\sum_{j=1}^{n} J(x_j;\hth)_{h}J(x_j;\hth)_{k},
\qquad
J(x_j;\hth)_h=\frac{\partial f(x_j;\bth)}{\partial\theta_h}\bigg|_{\bth=\hth},
\qquad
h,k=1,\ldots,p.
\label{eq:fim-emp}
\end{equation}

The FIM controls local distinguishability: expanding the Kullback-Leibler divergence
between two nearby members of the family, around $\hth$, gives
\begin{equation}
\KL\big(\hth\,\|\,\hth+\bdelta\big)
=\tfrac{1}{2}\,\bdelta^{\top} \Fhat_{[n]}\,\bdelta+\mathcal{O}(\|\bdelta\|^{3}).
\label{eq:kl}
\end{equation}
The FIM therefore measures how strongly a perturbation of the parameters
changes the predictive distribution. 

The $p\times p$ matrix $\Fhat_{[n]}$ is symmetric positive semi-definite
and we can write its spectral decomposition as
\begin{equation}
\Fhat_{[n]}=V\Lambda V^{\top},
\qquad
V=\left[v_1\cdots v_i\cdots v_p\right],
\qquad \Lambda=\mathrm{diag}(\lambda_1,\ldots,\lambda_i,\ldots,\lambda_p),
\label{eq:eig}
\end{equation}
with eigen-pairs $(\lambda_i,v_i)$ ordered by decreasing $\lambda_i$.
Directions with large $\lambda_i$ are locally very sensitive
and commonly named \emph{stiff},
while those with small $\lambda_i$ are weakly sensitive
and usually named \emph{sloppy} \citep{transtrum2015,Quinn2021}.

\subsection{Fisher distance to a parametric constraint}
\label{sec:prelim-wald}
 
Let $c:\Theta\to\R$ be a smooth scalar constraint whose zero set
$\{\bth\in\Theta\mid c(\bth)=0\}$ is the boundary of interest, and assume that $c$ can be
linearised and $\Fis_{[n]}(\theta)$ treated as constant in a neighbourhood of $\hth$.
The squared Fisher distance from $\hth$ to the boundary is obtained by
minimising $\bdelta^{\top}\Fhat_{[n]}\bdelta$ subject to the linearized constraint
$c(\hth)+\nabla c(\hth)^{\top}\bdelta=0$, which gives (more details in \cref{app:derivations})
\begin{equation}
\dF\;=\;
\frac{c(\hth)^{2}}
{\nabla c(\hth)^{\top}\,\Fhat_{[n]}^{-1}\,\nabla c(\hth)} .
\label{eq:fisherdistance}
\end{equation}
We read $\dF$ as a squared, curvature-normalised, signal-to-noise ratio
of the fitted solution against the hypothesis that the constraint holds.
Everything our model diagnosis in \cref{sec:taxonomy} will need,
besides the spectral decomposition in \cref{eq:eig},
is an instance of \cref{eq:fisherdistance} for a particular choice of $c$.

\paragraph{Calibration.}
In the interior of the parameter space and under standard regularity conditions,
we have
\begin{equation}
\hth\approx\mathcal{N}\!\big(\bth,\Fhat_{[n]}^{-1}\big)
\qquad
\mbox{and thus}
\qquad
c(\hth)\approx\mathcal{N}\!\big(c(\bth),{\nabla c(\hth)^{\top}}\Fhat_{[n]}^{-1}{\nabla c(\hth)}\big),
\label{eq:asympnorm}
\end{equation}
by the delta method, which makes \cref{eq:fisherdistance} the Wald statistic \citep{wald1943tests}
for $H_0:c(\bth)=0$; hence $\dF\approx\chi^{2}_{1}$ (more details in \cref{app:derivations}).
This $\chi^{2}_1$ reading of $\dF$ requires an interior, identifiable optimum,
a finite-variance limit, and a boundary that is smooth in the working coordinates.
Three of these can fail in our setting:
(i) for boundaries at infinity the normality of $c(\hth)$ is not realistic;
(ii) near non-identifiable configurations $\Fhat_{[n]}$ is ill-conditioned and the
variance estimate in the denominator of \cref{eq:fisherdistance} is unstable;
(iii) sample sizes are moderate, so the asymptotic regime cannot be fully trusted.
We therefore use \cref{eq:fisherdistance} only as a screening signal-to-noise ratio statistic:
a single, deliberately permissive, threshold $\tau$ admits a direction to the candidate set,
and the actual accept/reject decision is delegated to the predictive gate of \cref{sec:acceptance}.
Exactly non-identifiable directions are handled separately and never rely on \cref{eq:fisherdistance},
which becomes ill-defined in that limit.
The threshold used throughout is $\tau=16$, which we motivate by Chebyshev's inequality
rather than by Gaussian tails: for any standardized $Z$,
it holds that $\Pr(|Z|\ge k)\le k^{-2}$, so $\dF<\tau=k^{2}=16$ defines a broad compatibility region,
which holds at level at least $15/16$ without any distributional assumption. 

\section{Sensitivity is not enough: a taxonomy of parametric directions}
\label{sec:taxonomy}

Our diagnosis of a fitted symbolic model rests on two quantities.
The first is the local sensitivity $\lambda_i$ of \cref{eq:eig}, a~statement about the model's structure at $\hth$ that is tightly bound to the eigen-basis in which it is defined, but unrelated to any specific boundary.
The second is the boundary-compatibility statistic $\dF$ of \cref{eq:fisherdistance}, a statement about the fitted solution with respect to a specific boundary that is unrelated to any particular basis of the parameter space.

The local sensitivity is straightforward to compute, and it is standard.
We contend that the less trivial design choice is which boundary to test.
The eigen-hyperplane, $v_i^{\top}(\bth-\hth)=0$, may look like the natural candidate,
but it is only a local linear approximation to the model manifold,
and imposing a constraint on an eigen-direction is rarely a syntactic simplification of the expression;
it therefore does little for interpretability.
Testing constraints in the original parametric coordinates,
e.g., $\theta_k=0$, $\theta_k\to\pm\infty$, $\theta_k=\pm\theta_h$,
is instead something a computer algebra system can
impose directly, and it acts in the space in which the model is read.

Sensitivity and boundary compatibility are logically independent: $\lambda_i$ can be large while $\dF$ is small, and conversely.
It is exactly this independence that generates the taxonomy below, and it is what a criterion built on $\lambda_i$ alone, as in curvature-based model reduction, cannot see.
The two quantities do not enter symmetrically, however, and the asymmetry is what organises the classification.
A direction that is fully resolved by the data, i.e., it is stiff, can only be removed by a constraint that the fit already satisfies at a finite value: pushing a stiff coordinate towards an asymptotic limit would move the prediction, by construction.
Conversely, a coordinate sitting near an asymptotic regime has $\partial f/\partial\theta_k\approx 0$ on the observed design (the saturation limit), so it contributes negligibly to $\Fhat_{[n]}$ and it is necessarily sloppy. Stiff directions are therefore either used or removable to a finite value (generically $\theta_k=0$), whereas boundaries at infinity ($\theta_k\to\pm\infty$) belong to the sloppy case alone, and similarly for signed parametric identifications ($\theta_k=\pm\theta_h$). Furthermore, if $\lambda_i=0$, the direction is unidentifiable and the value of $\dF$ does not matter.

The classification below is intended as a phenomenological account of these mechanisms and, for expository purposes, associates each parametric coordinate with its best-aligned Fisher eigen-direction.
The algorithmic treatment in \cref{sec:method-tests} does not require such an association: it translates these cases into identifiability analysis and coordinate-level boundary screens, without assigning every parameter to a unique stiff or sloppy eigen-direction.

\subsection{Five canonical types}
\label{sec:tax-four}

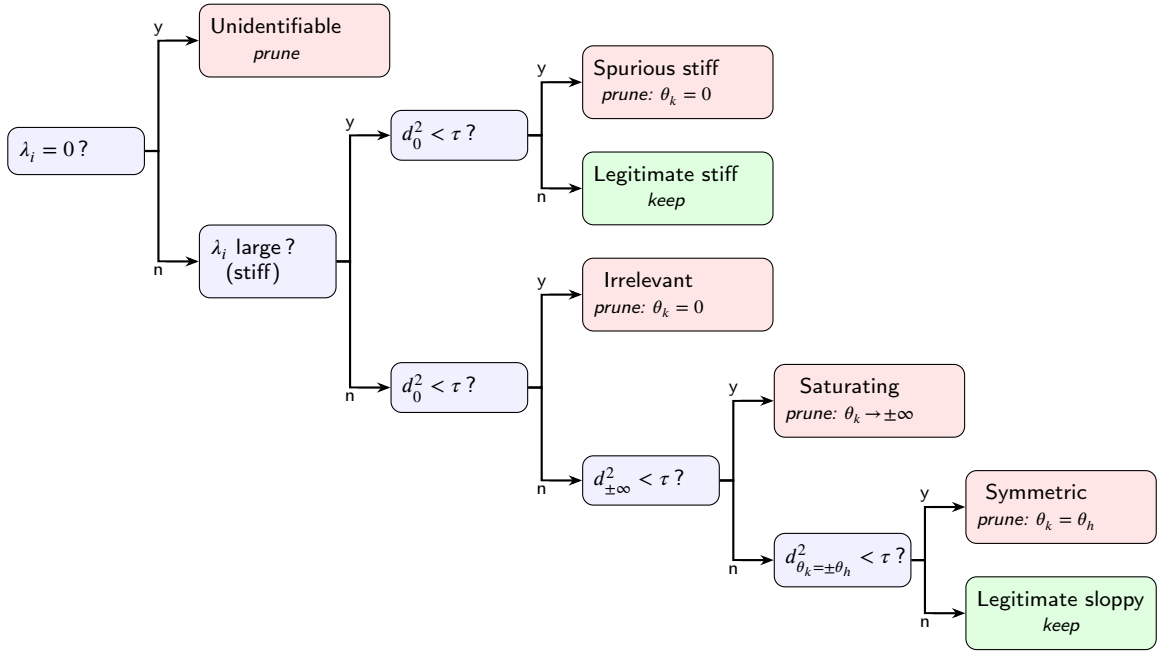
\begin{figure}[pos=!htbp]
\centering
\resizebox{\textwidth}{!}{%
\begin{forest}
forked edges, fork sep=10pt,
for tree={
    grow'=east, parent anchor=east, child anchor=west, anchor=west,
    draw, rounded corners, align=center,
    top color=blue!6, bottom color=blue!6,
    edge={-{Stealth[length=5pt]}, thick},
    l sep=20pt, s sep=12pt, inner sep=4pt, font=\footnotesize,
},
test/.style={text width=15mm},
prune/.style={top color=red!10, bottom color=red!10, text width=22mm},
keep/.style={top color=green!12, bottom color=green!12, text width=22mm},
yy/.style={edge label={node[pos=.5,above,el]{y}}},
nn/.style={edge label={node[pos=.5,below,el]{n}}},
[${\lambda_i=0}$\,?, test
    [Unidentifiable\\{\scriptsize\itshape prune}, prune, yy]
    [$\lambda_i$ large\,? \\(stiff), test, nn
        [$\dzero<\tau$\,?, test, yy
            [Spurious stiff\\{\scriptsize\itshape prune: ${\theta_k=0}$}, prune, yy]
            [Legitimate stiff\\{\scriptsize\itshape keep}, keep, nn]
        ]
        [$\dzero<\tau$\,?, test, nn
            [Irrelevant\\{\scriptsize\itshape prune: ${\theta_k=0}$}, prune, yy]
            [$\dinf<\tau$\,?, test, nn
                [Saturating\\{\scriptsize\itshape prune: $\theta_k\!\to\!\pm\infty$}, prune, yy]
                [$d^{2}_{\theta_k=\pm\theta_h}<\tau$\,?, test, nn
                    [Symmetric\\{\scriptsize\itshape prune: ${\theta_k=\theta_h}$}, prune, yy]
                    [Legitimate sloppy\\{\scriptsize\itshape keep}, keep, nn]
                ]
            ]
        ]
    ]
]
\end{forest}}
\caption{Taxonomy of parametric directions at the fitted solution, read as the
decision procedure of \cref{sec:method}. Internal nodes are the tests, edges
carry the answer (\textsf{y}/\textsf{n}); the two governing quantities are the
local sensitivity $\lambda_i$ and the boundary-compatibility statistic of
\cref{eq:fisherdistance}, evaluated on the coordinate constraints of
\cref{sec:method-tests} for the parameter $\theta_k$ dominating the direction.
Note the asymmetry argued in \cref{sec:taxonomy}: the stiff branch terminates in
the zero test alone, whereas the non-zero boundaries ($\theta_k\to\pm\infty$ and $\theta_k=\pm\theta_h$)
appear only in the sloppy branch.
Leaves marked \emph{prune} are the candidate targets of the  FGP reduction algorithm.}
\label{fig:taxonomy}
\end{figure}

Crossing the two quantities $\lambda_i$ and $\dF$ yields one of five cases,
with a further subdivision inside one of them.
Throughout, ``compatible with a boundary'' means $\dF<\tau$, assuming the boundary is specified by $c(\bth)=0$.
\Cref{fig:taxonomy} summarises the following classification
and the reduction pipeline described in \cref{sec:method-tests}.

\begin{enumerate}
\item \textbf{Unidentifiable.} Null $\lambda_i$, and $\dF$ does not matter.
The parameterisation is unidentifiable.
Typical examples include scale symmetries in rational expressions, where the numerator and denominator can be rescaled jointly, and multiplicative degeneracies such as $\theta_h\theta_k$, where only the product is identifiable and one parameter can be gauge-fixed, e.g.~by setting $\theta_k=1$.
More generally, structural non-identifiability arises whenever a model component depends on multiple parameters only through a lower-dimensional combination, as in nested affine transformations or nonlinear compositions such as $\theta_h e^{\theta_k}$.
Along a flat direction, the Fisher matrix is exactly singular, the statistic in \cref{eq:fisherdistance} degenerates, and the linearisation underlying it is in any case invalid.
This case is not uncommon for machine-generated expressions.

\item \textbf{Spurious stiff.} Large $\lambda_i$, but the fit is compatible with $\theta_k=0$.
The model is locally very sensitive along $v_i$, yet the estimator does not use the direction: a subexpression that could significantly affect predictions is switched off by a coefficient that the data place at zero.
Compatibility here is evidence for the boundary that survives a large inverse-variance factor, so these are among the safest and most profitable reductions available, and yet curvature-based reduction will never propose them, because $\lambda_i$ is large.
In randomly generated GP populations, they are among the most common sources of removable structure.

\item \textbf{Legitimate stiff.} Large $\lambda_i$, and the fit incompatible with $\theta_k=0$.
The direction is resolved by the data and used by the estimator: it carries the data-generating mechanism and must be preserved.

\item \textbf{Spurious sloppy.} Small $\lambda_i$, but the fit is compatible with a boundary.
This may be related to different mechanisms, which require different remedies to remove them.
    \begin{itemize}
    \item \emph{Irrelevant}: compatible with $\theta_k=0$. This is the same test as in the spurious-stiff case, applied at the opposite end of the spectrum: the direction is neither fully resolved by the data nor used by the estimator, and contributes nothing over the observed domain.
    \item \emph{Saturating}: compatible with $\theta_k\to\pm\infty$. The estimator has driven the coordinate into the asymptotic regime of a bounded nonlinearity (sigmoid, exponential tail), where the model output no longer responds. The correct reduction is the limit, which typically deletes an entire subtree; setting the same parameter to zero would instead leave a
    spurious constant residue.   
    \item \emph{Symmetric}:
    compatible with $\theta_k=\pm \theta_h$.
    This is an empirical analogue of structural non-identifiability: the two directions are identifiable, but the data do not distinguish between them; they can therefore be reduced to a single direction.
\end{itemize}

\item\textbf{Legitimate sloppy.}
Small $\lambda_i$, but the fit is incompatible with any given boundary (of the kind discussed above and again in \cref{sec:method-tests}).
The corresponding parameter $\theta_k$ is identifiable, contributes a finite and stable amount to the fit, and is compatible with no boundary.
Such directions are subleading, not spurious; removing them is a modelling decision (retaining only the leading contributions), not a correction.
\end{enumerate}

Extracting an emergent model then amounts to making the parameterisation identifiable,
removing the spurious directions and, optionally, the legitimate sloppy ones,
when only the leading mechanism is of interest.

The classification is reasonably sharp when the model is well-specified with respect to the data-generating mechanism, i.e., when the symbolic expression either contains the truth as a special case, possibly at a boundary, or approximates it well.
In this regime spurious directions correspond to degrees of freedom the estimator can suppress, either by shrinking coefficients to finite values or by pushing parameters to meaningful limits, while legitimate directions retain a non-negligible functional contribution.

Under severe misspecification, such as fitting a quadratic signal with a purely logarithmic model, no boundary or substructure approximates the truth.
Suppression mechanisms then become ambiguous: directions show mixed signatures, with neither clear saturation nor negligible contribution, and the separation between stiff and sloppy spurious modes blurs.
In this regime the search for an emergent model is intrinsically ill-posed, because the model to be reduced lacks a geometrically meaningful submanifold approximating the evidence.
The interpretability of the taxonomy should therefore be read as conditional on sufficient expressiveness of the  model.

\section{Fisher-Geometric Pruning}
\label{sec:method}

The FGP reduction operator takes a fitted symbolic model $(f,\hth)$ and returns a reduced symbolic model.
It is greedy in the space of admissible constraints and alternates three steps: \emph{propose}, \emph{realise}, and \emph{decide}.
In \emph{propose}, the Fisher spectrum diagnoses non-identifiable directions and boundary-compatibility statistics screen a small set of symbolically realisable constraints.
In \emph{realise}, each proposed constraint is applied to the expression, parameter-only subtrees are contracted, and the reduced candidate is refitted.
In \emph{decide}, realised candidates are tested by cross-validation; at most one move is accepted before the model is refitted and its geometry recomputed.
This separation ensures that neither the Fisher spectrum nor a boundary statistic certifies a reduction: geometry proposes, predictive validation decides.

The three stages are embedded in a fixed ordering of the reduction mechanisms.
Non-identifiable directions are treated first, because a singular $\Fhat_{[n]}$ makes the inverse-Fisher boundary statistics ill-defined; only after identifiability has been restored are the remaining boundary screens applied.

\subsection{Propose: geometric and boundary screening}
\label{sec:method-tests}

Sensitivity is defined in the Fisher eigen-basis, whereas a useful symbolic reduction must be expressed in the coordinates of the original expression.
We therefore never impose constraints such as $v_i^\top\bth=v_i^\top\hth$, nor do we construct a complete correspondence between eigen-directions and individual parameters.
Instead, the eigen-basis is only used to identify and represent the null subspace, from which we select the coordinates that best represent non-identifiable directions; all remaining parameters are assessed directly through their boundary statistics in the original coordinates.
This confines the mapping between the two bases to the one place where it is unavoidable and removes the need for a per-parameter notion of \emph{stiff} or \emph{sloppy} in the algorithm.

Three tests are used to check the constraint compatibility of identifiable parametric directions.
\Cref{tab:boundaries} collects the constraint function, its gradient and the resulting statistic; the derivation for the infinite saturation case is in \cref{app:derivations}.
Note that, while zeroing and infinite saturation test scalar limits, signed identification would require testing all pairs in the parametric coordinates.
To streamline computation, candidate pairs are pre-screened on the fitted magnitudes: a~pair $(h,k)$ is proposed with sign $s=\sgn(\hat\theta_h\hat\theta_k)$ when
\begin{equation}
\frac{2\,\big|\,|\hat\theta_h|-|\hat\theta_k|\,\big|}
     {|\hat\theta_h|+|\hat\theta_k|}\;<\;\eta,
\label{eq:identprescreen}
\end{equation}
where $\eta$ is a small relative tolerance ($\eta=10^{-2}$ in our experiments).
Note that the pre-screen is on magnitudes and that the sign $s$ is read off the fit (not tested in both variants).

\begin{table}[!htbp]
\centering\footnotesize
\caption{Boundary families for the identifiable directions, with their constraint functions and corresponding compatibility statistics. Here, $e_k$ denotes the $k$-th canonical basis vector.
All statistics are screened against the common threshold $\tau$ of \cref{sec:prelim-wald}.}

\label{tab:boundaries}
\renewcommand{\arraystretch}{1.6}
\setlength{\tabcolsep}{4pt}
\begin{tabular}{@{}p{3.0cm} p{3.0cm} p{2.5cm} p{3.5cm}@{}}
\toprule
\textbf{Family} & \textbf{Constraint $c(\bth)$} & \textbf{Gradient} {\boldmath$\nabla c(\bth)$} & \textbf{Statistic $\dF$} \\
\midrule
Zeroing
& $\theta_k$
& $e_k$
& $\dfrac{\hat\theta_k^{2}}{\big(\Fhat_{[n]}^{-1}\big)_{kk}}$ \\[15pt]
Infinite saturation
& $\phi_k\mp\frac{\pi}{2}$, $\phi_k=\arctan\theta_k$
& $e_k$ (in $\phi$)
& $\dfrac{\big(\arctan\hat\theta_k\mp\frac{\pi}{2}\big)^{2}\,
\big(1+\hat\theta_k^{2}\big)^{2}}{\big(\Fhat_{[n]}^{-1}\big)_{kk}}$ \\[15pt]
Signed identification
& $\theta_k\mp\theta_h$
& $e_k\mp e_h$
& $\dfrac{(\hat\theta_k\mp\hat\theta_h)^{2} }{(e_k\mp e_h)^{\top}\Fhat_{[n]}^{-1}(e_k\mp e_h)}$ \\
\bottomrule
\end{tabular}
\end{table}

\begin{itemize}
\item \textbf{Unidentifiable directions.}
These cannot be screened by \cref{eq:fisherdistance} as the statistic $\dF$ is meaningless when $\Fhat_{[n]}$ is  flat in the tested direction. They are therefore identified by a mask on the spectrum, followed by a nomination step.
The eigen-direction $i$ is flagged as unidentifiable when
\begin{equation}
\ln\lambda_i<\epsilon_\lambda
\qquad\text{or}\qquad
\max_h\ln\lambda_h-\ln\lambda_i>\Delta_\lambda
\;\;\text{and}\;\;\ln\lambda_i<0,
\label{eq:lowmask}
\end{equation}
i.e., when  its sensitivity is below either an absolute floor $\exp(\epsilon_\lambda)$
or a relative spectral gap of $\Delta_\lambda$ nats with respect to the stiffest direction ($\Delta_\lambda=25$ in our experiments).
The second clause makes the mask scale aware: an eigenvalue of $10^{-3}$ is unidentifiable in a problem with stiffest sensitivity $10^{9}$ and perfectly resolvable in one with stiffest sensitivity $10^{-1}$.

Let $w$ be the number of flagged eigenvalues, and let $V_{\mathrm{null}}\in\R^{p\times w}$ collect their eigenvectors. Because the symbolic constraints must ultimately be imposed on parameter coordinates, a $w$-dimensional null subspace requires the nomination of $w$ representative coordinates.
For this purpose, we define the participation of coordinate $k$ in the null subspace as the Euclidean squared norm of the projection of the $k$-th canonical basis vector on the null space:
\begin{equation}
u_k \;=\; \sum_{i\,:\,\text{flagged}} V_{ki}^{2}
\label{eq:participation}
\end{equation}
where $u_k\in[0,1]$, and $\sum_k u_k = w$, because the eigen-basis is orthonormal. Note that the eigen-basis of the null subspace is arbitrary, but $u_k$ is invariant under rotations of that basis and therefore provides a stable coordinate-level participation measure.

We select $w$ coordinates using a pivoted-QR-style greedy subset selection on $V_{\mathrm{null}}$ \citep{golub2013matrix}.
At each step, the coordinate with largest residual participation is selected; coordinates with comparable residual participation (within a factor $\rho$, here $\rho=0.9$) and collinear residual rows are recorded as equivalent representatives of the same null mechanism, and the row space is then deflated along the selected coordinate before the next step.
The collinearity condition prevents distinct null mechanisms with similar participation from being merged.

The coordinates in the resulting nominated set $\mathcal{U}$ are considered compatible with the scalar limits $0$, $1$, and~$\pm\infty$, together with the eligible signed identifications $\theta_k=\pm\theta_h$ given by \cref{eq:identprescreen}. The unit constraint is included because it provides a natural gauge fixing for common multiplicative redundancies without deleting the surviving term. Every parameter outside $\mathcal{U}$ is treated as identifiable and, crucially, is eligible for every boundary family listed below, gated only by its own $\dF$ in \cref{eq:fisherdistance}. 

\item \textbf{Identifiable directions.} The parameters not belonging to $\mathcal U$ are considered identifiable
and tested for the following limits.
\begin{enumerate}
\item \textbf{Zeroing.} The boundary $\theta_k=0$ is the standard elimination of a coefficient and every subtree that multiplies it. Constraint compatibility is tested with the Wald statistic.

\item \textbf{Saturating.} A limit $\theta_k\to\pm\infty$ cannot be tested directly,
so we compactify the coordinate:
\begin{equation}
\phi_k=\arctan\theta_k\in\left(-\tfrac{\pi}{2},\tfrac{\pi}{2}\right),
\qquad
\theta_k=\tan\phi_k,
\qquad
\frac{\partial\theta_k}{\partial\phi_k}=1+\theta_k^{2},
\label{eq:arctan}
\end{equation}
so that the boundary becomes the finite constraint $c(\phi)=\phi_k\pm\frac{\pi}{2}$.
The sign of the limit $s=\sgn(\hat\theta_k)$ is taken from the fit.
Recall that Gaussian calibration is not assumed: $\dF$ is read purely as a signal-to-noise ratio and screened by the Chebyshev-motivated threshold of \cref{sec:prelim-wald}.

\item \textbf{Symmetric.} We test $\theta_k=\pm\theta_h$ on eligible pairs.
These tests aim to catch data driven symmetries in the parameterisation, thus suggesting further lower-dimensional approximations, which the predictive gate is then asked to validate.
Only the pairs satisfying \cref{eq:identprescreen} together with the Wald screen in \cref{tab:boundaries} enter this candidate set.
\end{enumerate}
\end{itemize}

Note that our formulation excludes reductions that require genuine linear combinations of coefficients.
In return, every candidate reduction is an explicit symbolic substitution that can be implemented
by a computer algebra system and interpreted by a reader.

\subsection{Realise: constraint application and contraction}
\label{sec:method-contraction}
Let $\mathcal{C}=\{c_1,\dots,c_r\}$ denote the set of  candidate pruning moves for the current model, as determined by the propose step.
These may arise either from testing candidate limits for unidentifiable parameters $\theta_k\in\mathcal{U}$ or from candidate constraints associated with identifiable directions.
Imposing a constraint $c_m$ produces a candidate model
\[
f_m(\,\cdot\,;\bth_{c_m}),
\]
where $\bth_{c_m}$ denotes the parameter components remaining after substitution and
algebraic contraction.

The contraction step removes residual parametric redundancy and restores a canonical parameterisation.
After substitution, every maximal subtree depending only on parameters is collapsed and reparameterised as a single constant.
For instance, $\theta_1\theta_2+\log(\theta_3)\mapsto\theta$, where the new parameter is warm-started at the numerical value of the collapsed subtree under the current fit. Limits at zero or infinity may also annihilate an entire subtree. For instance, $e^{\theta_1x+\theta_2}\to 0$ as $\theta_1\to-\infty$ on a domain with $x>0$, thereby removing all parameters contained in that subtree.
Consequently,
\[
|\bth_{c_m}|\leq|\bth|-1,
\]
i.e., in general, a single pruning move may reduce the parameter dimension by more than one, and different constraints applied to the same model may produce candidates of different dimensions.

Our symbolic manipulation is deliberately restricted to purely parametric subtrees and does not simplify expressions involving covariates. A substitution may therefore introduce removable singularities at the covariate level. For example, $\frac{x}
     {\theta_1+\theta_2x+\theta_3x^2}
\;\longmapsto\;
\frac{x}{\theta_2x+\theta_3x^2}$,
which is formally undefined at $x=0$ and may cause numerical instability when evaluated,
even though a common factor could be cancelled.
Such cases could be handled by an additional symbolic-simplification step.
We deliberately omit this step, in the present implementation,
so that the empirical effects of the proposed model-reduction procedure remain distinguishable
from those of standard semantics-preserving algebraic rewriting.

\subsection{Decide: predictive admissibility and parsimony tie break}
\label{sec:acceptance}
The standard dimension-penalised criteria commonly used in SR, including AIC, BIC, and MDL, differ in their inferential targets, i.e., expected predictive risk, marginal likelihood, and code length, respectively, but their usual formulations face three difficulties in symbolic pruning.

First, pruning deliberately traverses non-regular configurations.
Standard penalties are derived for locally regular models with an identifiable optimum in the interior, non-singular Fisher information, and a quadratic log-likelihood approximation \citep{akaike1974new,schwarz1978estimating,kass1995bayes,grunwald2007mdl}.
These conditions can fail at the boundaries, singularities, and non-identifiable parameterisations encountered during reduction, where the nominal dimension does not need to provide the appropriate complexity correction \citep{watanabe2009algebraic,watanabe2013widely}.

Second, SR pruning must work without assuming that the data-generating mechanism belongs to the symbolic hypothesis space, much less to the finite population returned by the search. Under misspecification, the usual AIC correction is no longer generally valid, while the marginal-likelihood interpretation underlying BIC may be poorly aligned with predictive performance \citep{bernardo1994bayesian,grunwald2017inconsistency,vehtari2017practical}.

Third, parameter count does not capture the functional mechanisms through which symbolic expressions can overfit. A relatively low-dimensional expression may still achieve an excellent in-sample fit through a finely tuned cancellation, a near-pole, or another unstable algebraic configuration that does not survive re-estimation.

These considerations motivate a more operational acceptance rule. Rather than asking whether a reduced expression receives stronger in-sample evidential support, we ask whether lower complexity can be achieved without degrading out-of-sample predictive performance. Each candidate reduction is therefore refitted and evaluated by cross-validation. Only among candidates that cross-validation cannot meaningfully separate do we use BIC as a parsimony-based tie-break. This two stage procedure, separating predictive admissibility from parsimony-based tie-break is described below.

\paragraph{Step 1: predictive admissibility.}
Let $\{F_1,\dots,F_K\}$ be a fixed random partition of the sample in $K$ folds,
and $\hat f^{(-k)}$ the model refitted on all folds but the $k$-th.
The cross-validated risk and its across-fold standard error are
\begin{equation}
\Rcv(f)=\frac{1}{K}\sum_{k=1}^{K}\frac{1}{|F_k|}\sum_{j\in F_k}\big(y_j-\hat f^{(-k)}(x_j)\big)^{2}
\quad
\mbox{and}
\quad
\scv(f)=\frac{1}{\sqrt{K}}\,
\mathrm{sd}\left(\frac{1}{|F_k|}\sum_{j\in F_k}\big(y_j-\hat f^{(-k)}(x_j)\big)^{2}\right)_{k=1}^K.
\label{eq:cv}
\end{equation}
A reduction $f\mapsto f_m$ is \emph{admissible} when it does not degrade the cross-validated risk
by more than a relative tolerance:
\begin{equation}
r(f_m,f)\;=\;\frac{\Rcv(f_m)-\Rcv(f)}{\Rcv(f)}\;\le\;\beta,
\label{eq:cv-accept}
\end{equation}
where $\beta>0$ is the most important knob of our method;
it allows to tune how much predictive fidelity one is willing to trade for structural simplicity. 

\paragraph{Step 2: parsimony tie-break.}
Cross-validation has finite resolution. Writing $f_{\star}$ for the admissible candidate with the smallest $\Rcv$,
the candidates it cannot separate form the set
\begin{equation}
\mathcal{E}=\Big\{f_m \;\Big|\; r(f_m,f)\le\beta,\;\;
\Rcv(f_m)-\Rcv(f_{\star})\le\kappa\,\scv(f_{\star})\Big\},
\label{eq:equivset}
\end{equation}
using a standard-error-type rule with $\kappa$ of order one
\citep{breiman1984classification,hastie2009elements}.
Admissible reductions can have different dimensions because a single symbolic constraint may remove several parameters after contraction.
We therefore use model dimension, only after predictive admissibility has been established, to resolve candidates that cross-validation cannot separate.
The tie is broken by a dimension-aware criterion:
\begin{equation}
\bar f=\argmin_{f_m\in\mathcal{E}}\;\mathrm{BIC}(f_m),
\qquad
\mathrm{BIC}(f_m)=n\Big(1+\ln\big(2\pi\,\widehat{\mathrm{MSE}}(f_m)\big)\Big)
+|\bth_{c_m}|\ln n ,
\label{eq:bic}
\end{equation}
where $\widehat{\mathrm{MSE}}$ is the in-sample mean squared error at the
full-data MLE of the reduced model and the noise level is not counted as a free parameter. 

The role of BIC is thus deliberately restricted: it never decides admissibility
and never certifies that a reduced expression is true.
It is consulted only after the predictive gate has already discarded the fragile candidates,
at which point a parameter count is a transparent and adequate preference for the simpler symbolic representation.

\subsection{Complete algorithm and computational cost}
\label{sec:method-algorithm}

Having defined candidate generation, symbolic realisation, and predictive acceptance separately, we now assemble the three stages into the complete greedy reduction procedure. \Cref{alg:main} gives the overall orchestration.

The algorithm calls four helper routines whose mathematical ingredients have been defined above. \textsc{Geometry} (\Cref{alg:geom}) forms the empirical Fisher matrix, computes its spectrum and pseudoinverse, nominates the null-space coordinate set $\mathcal{U}$ through \cref{eq:lowmask,eq:participation}, and evaluates the boundary-compatibility statistics of \cref{tab:boundaries}. \textsc{Screen} constructs the mode-specific candidate constraints from $\mathcal{U}$ and the corresponding boundary screens. \textsc{BestReduction} applies the admissible constraints for one candidate, contracts and refits the resulting expressions, and evaluates their CV risk and BIC. Finally, \textsc{Select} implements the two-stage rule of \cref{sec:acceptance}: predictive admissibility first, followed by BIC tie-breaking within the CV-equivalent set. Detailed pseudocode of the \textsc{Screen}, \textsc{BestReduction}, and \textsc{Select} subroutines is reported in appendix \ref{app:routines}.

Starting from the fitted expression, the geometry record $\mathcal{G}$ is computed at initialization. The four reduction modes
\begin{equation}
\textsc{unidentifiable}\;\to\;\textsc{zeroing}\;\to\;
\textsc{saturating}\;\to\;\textsc{symmetric}
\label{eq:modeorder}
\end{equation}
are then swept in this fixed order, with each mode iterated to exhaustion before the next is entered. The complete sweep is repeated until a full pass accepts no move. After every accepted reduction, the model is refitted and $\mathcal{G}$ is recomputed from scratch, because the optimum of the reduced model, and hence its local geometry, is not obtained by simply restricting that of its parent. Within each mode, \textsc{Screen} determines the eligible candidate constraints and \textsc{BestReduction} selects the representative reduction associated with each candidate parameter. \textsc{Select} is then applied a second time across these representatives to choose at most one move for the current iteration. Thus, the same predictive-admissibility and parsimony rule is used both when choosing among alternative boundaries for one parameter and when choosing among candidate parameters within a mode.

\begin{algorithm}[t]
\caption{\textsc{FisherGeometricPruning}  -- main loop}
\label{alg:main}
\begin{algorithmic}[1]
\Require \bf~expression $f$, data $\D$;
tolerances $\beta,\kappa,\tau,\epsilon_\lambda,\Delta_\lambda,\eta$; folds $K$
\Ensure reduced expression $f$
\State \bf~$\hth\gets\textsc{Fit}(f,\D)$
    \Comment{multi-start bounded least squares, ridge continuation}

\State \bf~$(c,s_c)\gets\textsc{CvRisk}(f,\hth,\D)$;\quad $B\gets\textsc{Bic}(f,\hth,\D)$
\State \bf~$\mathcal{G}\gets\textsc{Geometry}(f,\hth,\D)$ \Comment{Fisher spectrum, null set, boundary statistics}
\Repeat
  \State $\mathit{progress}\gets\textsc{false}$
  \For{$m\in\langle\textsc{unidentifiable},\textsc{zeroing},\textsc{saturating},\textsc{symmetric}\rangle$}
      
     \Loop
        \State $\mathcal{K}\gets\textsc{Screen}(\mathcal{G},m)$ \Comment{mode-specific admissible coordinate constraints}
        \If{$\mathcal{K}=\varnothing$} \textbf{break} \EndIf
        \State $\mathcal{M}\gets\{\textsc{BestReduction}(k,\mathcal{G},f,\hth,\D):k\in\mathcal{K}\}$
               \Comment{substitute, contract, refit, CV-evaluate}
        \State $(\mathrm{move}^{\star},\mathit{ok})\gets\textsc{Select}(\mathcal{M},c)$
               \Comment{CV admissibility, then BIC within equivalence band}
        \If{$\neg\,\mathit{ok}$} \textbf{break} \EndIf

        \State \bf~apply $\mathrm{move}^{\star}$: $f\gets$ contracted reduced model
        \State \bf~$\hth\gets\textsc{Fit}(f,\D)$ warm-started at the screening solution
               \Comment{tight tolerances}
        \State $(c,s_c)\gets\textsc{CvRisk}(\cdot)$;\ $B\gets\textsc{Bic}(\cdot)$;\
               $\mathcal{G}\gets\textsc{Geometry}(\cdot)$;\ $\mathit{progress}\gets\textsc{true}$
     \EndLoop
  \EndFor\Until{$\neg\,\mathit{progress}$}
\State \Return $f$
\end{algorithmic}
\end{algorithm}

\begin{algorithm}[t]
\caption{\textsc{Geometry} -- Fisher spectrum, load map and boundary distances}
\label{alg:geom}
\begin{algorithmic}[1]
\Require \bf~$f,\hth,\D$
\Ensure geometry record $\mathcal{G}$
\State \bf~$J_{jh}\gets\partial f(x_j;\hth)/\partial\theta_h$,
       $j=1,\ldots,n$, $h=1,\dots,p$;\quad
       $\hat\sigma^{2}\gets\frac1n\|y-f(x;\hth)\|^{2}$
\State \bf~$\Fhat_{[n]}\gets \frac{1}{\hat\sigma^{2}} J^{\top}J$;\quad
       $\Fhat_{[n]}^{-1}\gets\mathrm{pinv}(\Fhat_{[n]})$
\State \bf~$(\lambda,V)\gets\mathrm{eigh}\big(\Fhat_{[n]}\big)$;\quad
       $\mathcal{U}\gets\textsc{NominateUnidentifiable}(\lambda,V)$ 
       \Comment{\cref{eq:lowmask,eq:participation}}
\For{\bf~$k=1,\dots,p\;\cap \theta_k\notin\mathcal U $}
  \State $\dF[k]\gets\big(d^{2}_{0},d^{2}_{+\infty},d^{2}_{-\infty}\big)(k)$
         \Comment{\cref{tab:boundaries}}
\EndFor
\State \bf~$\mathcal{P}\gets$ pairs $(h,k)$ passing \cref{eq:identprescreen},
       with $d^{2}_{\mathrm{sgn}}(h,k)$ from \cref{tab:boundaries}

\State \Return \bf~$\mathcal{G}=\{\Fhat_{[n]}^{-1},\lambda,V,\dF,\mathcal{P}\}$
\end{algorithmic}
\end{algorithm}

\paragraph{Numerical reliability of the fits.}
The geometric diagnosis depends on the quality of the fitted constants:
$\Fhat_{[n]}$, its spectrum, the nominated null-space coordinates, and all
boundary statistics in \cref{tab:boundaries} are evaluated at the fitted
solution. This dependence is not specific to our method. Any SR evaluation
that scores a numerically calibrated expression, for example through a
likelihood or likelihood-penalised fitness, can be distorted by an imperfect
coefficient fit \citep{kommenda2020parameter,cranmer2023pysr}.
What is specific to the present pruning setting is that the consequences need
not remain local: an inaccurate fit can alter the geometric diagnosis that
selects the next structural reduction, and the resulting error can therefore
propagate along the greedy trajectory.

Two failure modes are particularly relevant. If the optimiser stops short of
the optimum, a residual gradient can make a direction appear artificially
flat and send a legitimate parameter into the null-space candidate set. If the
fitted optimum is strongly ill-conditioned, as can occur for near-pole
configurations, the inverse-Fisher quantities entering the boundary statistics
become numerically unstable. We therefore use a two-tier fitting strategy:
candidate reductions and CV folds are evaluated with cheaper warm-started
screening fits, whereas every accepted reduction is refitted with tight
multi-start bound-constrained least squares before its geometry is recomputed.
Thus, no new geometric diagnosis is propagated from a screening-quality fit.
The complete optimisation protocol, including ridge continuation, restart
strategy, conditioning guards, bounds, and convergence tolerances, is given in
\cref{app:numerics}.

\paragraph{Computational cost.}
Let $p$ be the current number of parameters, $n$ the sample size, $K$ the number of folds,
and $C_{\mathrm{fit}}(n,p)$ the cost of one nonlinear least-squares solve.
A call to \textsc{Geometry} requires $O(np^{2})$ operations to form $J^{\top}J$
and $O(p^{3})$ operations for the spectral decomposition and pseudo-inverse.
Once $\Fhat_{[n]}^{-1}$ is available, the scalar boundary statistics require $O(p)$ operations, while screening all signed-identification pairs requires $O(p^{2})$ operations.

Let $R_p$ denote the number of realised candidate reductions at dimension $p$.
Each candidate requires one full-data screening fit and $K$ fold-specific fits,
and therefore costs $(K+1)C_{\mathrm{fit}}(n,p)$.
One iteration consequently has cost $O\!\left(np^{2}+p^{3}+(K+1)R_p C_{\mathrm{fit}}(n,p)\right)$.
For the scalar boundary modes, $R_p=O(p)$,
whereas signed identification may produce $R_p=O(p^{2})$ candidates in the worst case.
Across all iterations, since every accepted reduction removes at least one parameter,
at most $p$ moves can be accepted. Summing over the decreasing model dimensions gives
\begin{equation} 
O\!\left(
np^{3}+p^{4}+(K+1)p^{3}C_{\mathrm{fit}}(n,p)
\right)
\label{eq:cost}
\end{equation}
as a conservative worst-case bound when all signed pairs survive the pre-screen.
Candidate and fold evaluations are independent and are parallelised in the implementation.

\section{Experimental setup}
\label{sec:setup}

\subsection{Case studies}
We evaluate the FGP reduction operator at two complementary levels: controlled diagnostic problems isolate individual reduction mechanisms under known $M$-closed and $M$-open conditions; final populations produced by SR test whether the same mechanisms remain useful on heterogeneous, machine-generated expressions.
In this setting, we call a problem $M$-closed when the data-generating function can be represented by the candidate parametric family for some parameter configuration, possibly attained only as a boundary limit (such as $\theta_k\to\pm\infty$); when no such configuration or limiting parameterisation exists, we call the problem $M$-open.
Thus, in the terminology of parametric modelling, the former correspond to well-specified problems (up to closure of the model family), whereas the latter are genuinely misspecified.

\paragraph{\textbf{Controlled diagnostic problems.}}

We construct ten controlled problems for which the overparameterised candidate, the data-generating function, and the intended reduction mechanism are known. 

The suite spans regular, near-singular, singular, and $M$-open regimes, while also covering the main reduction mechanisms considered here. The ten problems are: nested sine-quadratic (\#1), a regular case combining removable spurious terms with a boundary reduction; one- and two-dimensional expansions (\#2-\#3), which test reduction under misspecification; near-alias Fourier (\#4), which probes weak identifiability caused by competing representations; Bessel approximation (\#5), an $M$-open case with unstable rational behaviour; rational function (\#6), competing saturations (\#7), and separable 2D (\#8), which introduce exact non-identifiabilities associated with overparameterised rational structures; buried transcendentals (\#9), which requires asymptotic removal of terms that cannot be zeroed directly; and tanh gauge (\#10), which requires signed parameter identification. Problems~\#5, \#6, and~\#8 additionally admit pole configurations inside or near the sampling domain. All problems use a training sample of $n=100$ observations and are evaluated at two levels of additive Gaussian noise. Full definitions, geometric classification, and expected reduction paths are reported in \cref{app:cases}.

\paragraph{\textbf{Pareto fronts from GPSR.}}
This setting probes the intended use of the method: post-hoc reduction of heterogeneous, machine-generated symbolic models that may contain redundant structure and sampling-specific overfitting.
Candidate expressions are generated with multi-objective GP using a non-dominated sorting algorithm (as implemented in Operon~\cite{burlacu2020operon}), using negative log-likelihood and node count as the two search objectives to be minimized.
For each experimental condition, we retain all individuals belonging to the first non-dominated front of the final population in the $(\mathrm{NLL},\#\mathrm{nodes})$ plane.
Thus, no single model is selected before pruning: the method is applied independently to all expressions representing the final accuracy-complexity trade-off.
The set of expressions deliberately contains underfit and overfit expressions to test the capabilities of FGP.

Each expression is parsed into symbolic form, its numerical constants are re-extracted as free parameters and initialised at the values returned by tree-based GP, and the reduction procedure of \cref{alg:main} is then applied.

The GPSR runs use the same algorithms and configuration as reported in~\cite{kronberger2026guiding}, that is, $200$ generations with a population of $1000$ individuals and a maximum length limit of 100 nodes.
We use the function set $\{+, -, \times, \div, x^2, \sin, \cos, \exp, \log |x|, \sqrt{|x|}, |x|^y\}$ with abs-protected operators, and terminal set \{<var>, <param>, <var> $\times$ <param>\}.
Parents are selected with tournament selection from a group of two random individuals based on Pareto ranks with crowding-distance as a tie break as in NSGA-II~\cite{Deb2002}.
Evolutionary variation operators are subtree-crossover (probability 90~\%) and node mutations for the remaining individuals (5~\% per-node mutation probability).

We consider two ten-dimensional synthetic regression problems based on
\citet{friedman1991mars}. In both cases, $x_h\sim\mathcal{U}[0,1]$, independently for $h=1,\ldots,10$, while only the first five variables affect the target; the remaining five therefore act as irrelevant inputs.
The first benchmark, denoted \emph{Friedman}, is
\begin{equation}
f(x)=10\sin(\pi x_1x_2)+20(x_3-0.5)^2+10x_4+5x_5+\varepsilon,
\qquad
\varepsilon\sim\mathcal{N}(0,\sigma^2),
\label{eq:friedman}
\end{equation}
and combines a nonlinear interaction, a quadratic effect, and two linear effects. It is evaluated at fixed $n=100$ with increasing noise,
$\sigma\in\{0.5,1.0,13.6\}$, a representative subset of those considered in \cite{kronberger2026guiding}.

The second benchmark, denoted \emph{Friedman additive}, uses the same input distribution but replaces the interaction structure with additive nonlinear effects:
\begin{equation}
f(x)=0.1e^{4x_1}
+\frac{4}{1+e^{-20(x_2-0.5)}}
+3x_3+2x_4+x_5+\varepsilon,
\qquad
\varepsilon\sim\mathcal{N}(0,\sigma^2).
\label{eq:friedman-additive}
\end{equation}
It combines exponential and sigmoidal effects with three linear terms.
Here $\sigma=1$ is fixed while the training sample size varies as $n\in\{50,100,1000\}$.

The two series therefore probe complementary axes: increasing noise at fixed sample size and increasing sample size at fixed noise.
Predictive quality is measured on an independent test set of $1000$ points drawn from the same input distribution and evaluated against the noiseless generating function.
Test MSE therefore measures agreement with the underlying signal, while cross-validation used by the reduction procedure is computed exclusively on the training data.

\subsection{Benchmarking against baselines and ablations}
\label{sec:setup-ablation}

Because the object of study is a reduction operator rather than a search algorithm, the informative comparison is against reduction operators that share its machinery and differ in one design choice.
For this reason, we consider a set of four methods representing existing approaches but also ablation variants.
The aim is to compare our approach with these competitors starting from the same fitted model in the fairest possible way to isolate the pruning approach.
For this reason, every shared component, such as fitting routine, tolerances, Fisher computation, contraction, refitting, is held fixed, so that the comparison is paired at the level of individual expressions.
\Cref{tab:ablation-methods} summarises the four alternatives.

\begin{table}[!htbp]
\centering\footnotesize
\caption{Alternative reduction operators. All variants share the fitting,
contraction, and refitting machinery of \cref{alg:main}.}
\label{tab:ablation-methods}
\begin{tabular}{@{}p{2.1cm} p{4.0cm} p{3.2cm} p{4.4cm}@{}}
\toprule
\textbf{Variant} & \textbf{Candidate selection} & \textbf{Allowed moves} & \textbf{Acceptance} \\
\midrule
\texttt{FGP} (ours)
& staged according to \cref{eq:modeorder}; null-space nomination followed by boundary screens
& $0,1,\pm\infty$, signed identification
& CV admissibility \cref{eq:cv-accept} + BIC tie-break \\[30pt]

\texttt{sloppy\_only}
& parameter with largest loading on the current smallest-eigenvalue direction
& same boundary families as \texttt{FGP} for the selected parameter
& CV admissibility + BIC tie-break \\[30pt]

\texttt{zero\_only}
& all parameters satisfying $d_0^2<\tau$ null-space/eigenvalue gating disabled
& $\theta_k\to0$ only
& CV admissibility + BIC tie-break \\[30pt]

\texttt{magnitude}
& smallest $|\hat\theta_k|$ below
$\tau_{\mathrm{mag}}=f_{\mathrm{mag}}\max_h|\hat\theta_h|$
& $\theta_k\to0$ only
& threshold only; no geometric or predictive test \\[30pt]

\texttt{bic\_only}
& identical to \texttt{FGP}
& identical to \texttt{FGP}
& $\Delta\mathrm{BIC}<B_{\mathrm{acc}}$; CV not used
\\
\bottomrule
\end{tabular}
\end{table}

\paragraph{\textbf{Sensitivity as the only signal} (\texttt{sloppy\_only}).}
This variant isolates the principle that reduction should follow the least-sensitive Fisher direction, as in sloppiness-based model-reduction approaches \citep{transtrum2014,transtrum2015,berman2023,howard2025}.
At each iteration, the Fisher spectrum is recomputed and the parameter with the largest loading on the smallest-eigenvalue direction is selected; its admissible boundary moves are then evaluated using the same predictive rule as our method.
This is not intended as a literal implementation of MBAM or Bayesian renormalisation, but as an ablation representing their shared emphasis on weakly resolved directions.

\paragraph{\textbf{Reduction restricted to the zero boundary} (\texttt{zero\_only}).}
This variant retains the same boundary-compatibility statistic and predictive acceptance rule as \texttt{FGP}, but permits only $\theta_k=0$ reductions, thus following the conventional coefficient-elimination setting considered by zero-pruning methods, including recent Fisher-geometric approaches \citep{berman2026optimal}.
Unlike these approaches, however, \texttt{zero\_only} tests compatibility with the zero boundary through $d_0^2$ and retains the predictive acceptance rule of FGP.
The eigenvalue/null-space gating is disabled so that every parameter can be considered directly through $d_0^2$.
It therefore isolates the contribution of the non-zero reduction mechanisms: unit constraints (for unidentifiable parameters), asymptotic limits, and signed parameter identifications.

\paragraph{\textbf{Coefficient-magnitude pruning} (\texttt{magnitude}).}
This variant provides a statistics-free coefficient-pruning baseline, in the spirit of magnitude pruning \citep{han2015learning,blalock2020state}.
At each iteration, the smallest coefficient below the relative threshold $\tau_{\mathrm{mag}}=f_{\mathrm{mag}}\max_h|\hat\theta_h|$ is set to zero and the model is refitted, until no coefficient remains below threshold.
No Fisher information, boundary statistic, or predictive acceptance test is used.
This comparison therefore tests whether raw coefficient magnitude alone is a sufficient pruning signal despite its dependence on parameterisation and scale.

\paragraph{\textbf{BIC-based acceptance} (\texttt{bic\_only}).} 
This variant retains the complete candidate-generation and symbolic-reduction machinery of \texttt{FGP} and changes only the decision rule. Specifically, candidate moves are ranked by $\Delta\mathrm{BIC}=\mathrm{BIC}(f_m)-\mathrm{BIC}(f)$ and the best move is accepted when $\Delta\mathrm{BIC}<B_{\mathrm{acc}}$; cross-validation is not consulted during the reduction trajectory.
This directly isolates the role of predictive validation in \cref{sec:acceptance}.

\subsection{Hyperparameter settings}
\label{sec:setup-hyper}
Unless stated otherwise, all experiments use the same reference configuration, which is fixed across problems and experimental settings rather than tuned case by case.
The reduction operator uses predictive tolerance $\beta=0.1$, CV equivalence-band parameter $\kappa=2$, and boundary-screening threshold $\tau=16$.
Numerical non-identifiability is detected using $\epsilon_\lambda=-10$ and a spectral-gap threshold $\Delta_\lambda=25$ on the log-eigenvalue scale.
Signed-identification candidates are pre-screened with $\eta=0.01$, while gauge representatives use $\rho=0.9$.
Predictive risk is estimated by $8$-fold cross validation. For each dataset, a fixed fold partition generated with seed $0$ is reused across candidate comparisons and across all variants that invoke cross-validation, ensuring paired evaluation.

All variants use the same numerical fitting protocol of \cref{app:numerics} whenever a fit or refit is required. In particular, the same optimisation procedure is used for the initial expression and for reduced candidates; the complete specification of initialisation, ridge continuation, parameter bounds, screening and diagnostic tolerances, and convergence guards is reported there.
For the ablation variants, magnitude pruning uses $f_{\mathrm{mag}}=0.01$, while \texttt{bic\_only} uses the acceptance band $B_{\mathrm{acc}}=3$.

\subsection{Metrics and statistical protocol}
\label{sec:setup-metrics}

The evaluation addresses three primary objectives: functional recovery,
predictive accuracy, and model complexity. Pareto dominance is used only as a
joint summary of the latter two objectives.

\paragraph{Replication and pairing.}
Each controlled diagnostic configuration is replicated over $30$ independent seeds, varying both the sampled dataset and the numerical optimisation initialisation.
All reduction operators are evaluated on the same seeds, so comparisons are paired at the replication level.

For the GPSR experiments, each noise/sample-size condition is represented by the first non-dominated front of a separate evolutionary run.
Each reduction operator is applied once to every expression on that front, and the original expression provides the paired unit across methods.
Repeating the pruning procedure on the same expression would primarily measure numerical optimisation variability rather than provide an independent model instance; we therefore compare methods expression-wise within each Pareto front.

\paragraph{Functional recovery.} 
For controlled problems in which the data-generating function belongs to the candidate family or its closure, we assess whether pruning recovers the generating model.
A syntactic criterion is inappropriate because equivalent functions may admit multiple algebraic representations, and some problems exhibit gauge freedoms. We therefore define recovery functionally.
A returned expression $g(x;\varphi)$ is said to recover $f^{\star}$ if: 
\begin{enumerate} 
\item[(i)] \emph{sufficiency}: after refitting $\varphi$ to the noiseless generating function on the design, the relative RMSE is below a numerical noise level (here set to $10^{-8}$); 
\item[(ii)] \emph{parsimony}: $|\varphi|\le p_{\star}$, where $p_{\star}$ is the number of independent constants in $f^{\star}$ after gauge fixing. 
\end{enumerate} 
The first condition ensures functional equivalence independent of parameterisation; the second prevents over-parameterised expressions from being counted as recovered.
We report both sufficiency and recovery rates, as their difference quantifies residual redundancy.
For all controlled set experiments belonging to the $M$-closed universe (seven problems at two noise levels), we compare our method (\texttt{FGP}) with each competitor via the net recovery rate 
\begin{equation} 
\Delta_{\mathrm{rec}} = \widehat{\Pr}[\text{\texttt{FGP} recovers}] - \widehat{\Pr}[\text{competitor recovers}] \in[-1,1], 
\end{equation} 
estimated over the 30 paired seeds. Configuration-level differences are tested against zero using a two-sided Wilcoxon signed-rank test with Holm correction across the four competitors. This aggregation respects the nesting of seeds within configurations. Additionally, we report discordant counts ($b/c$) as descriptive support only. 

\paragraph{Predictive accuracy and model complexity.} Accuracy is measured by $\mathrm{MSE}_{\star}$ against the noiseless truth. Complexity is measured by node count or the number of free parameters.

We first analyse accuracy and complexity separately.
Since each reflects only one axis of the underlying trade-off, neither is sufficient alone to characterise overall performance but they are the most interpretable quantities.
In the controlled experiments, following standard multiple-dataset methodology \citep{demsar2006statistical,garcia2010advanced}, each of the $N_c=20$ configurations is treated as one observation, summarised by the median over 30 seeds.  

In the SR-population experiments, each Pareto front is first restricted to a single common active subset containing all expressions for which at least one of the evaluated reduction operators performs at least one pruning move.
This subset is defined once per population and is then used for all pairwise method comparisons.
The restriction removes expressions that remain unchanged under every operator, which would otherwise contribute uninformative zero differences, particularly on the low-complexity region of the first Pareto front. 

We report average ranks (rank 1 = best), median paired difference, and compare \texttt{FGP} against each competitor using paired Wilcoxon signed-rank tests.
Holm correction is applied separately for each objective. Effect sizes are reported via rank-biserial correlation. 

Pareto dominance is used as a deterministic summary of the two marginal objectives to clarify potential ties or unclear single objective results: for example, two operators can match on each objective separately while showing a clear hierarchy on the pair, because one purchases parsimony with accuracy or the reverse.
Within a paired unit, $A$~Pareto-dominates $B$ if it is no worse in both complexity and $\mathrm{MSE}_{\star}$ and strictly better in at least one.
This relation is scale-invariant and therefore robust to the large dynamic range of $\mathrm{MSE}_{\star}$.
For each configuration and competitor, we compute the net dominance rate 
\begin{equation} 
\label{eq:netdom}
\Delta_{\mathrm{dom}} = \widehat{\Pr}[\text{\texttt{FGP} dominates}] - \widehat{\Pr}[\text{competitor dominates}] \in[-1,1], 
\end{equation} 
over the different pairs.
These configuration-level summaries are tested against zero using a Wilcoxon signed-rank test with Holm correction across competitors.
This analysis is a post-processing of the two primary objectives and should be interpreted jointly with the marginal tests. All tests are two-sided with familywise significance level $5\%$.

\section{Results}
\label{sec:results}

The experimental results support three main conclusions. First, our method successfully prunes symbolic models across both ends of the Fisher spectrum and through both finite and asymptotic reduction limits (\cref{fig:case-bessel}). Accepted reductions decrease structural complexity while preserving or improving agreement with the noiseless generating function, with this effect particularly clear on the evolved GPSR expressions and predictive improvements becoming more frequent as noise or sample size increases (\cref{fig:population-full}). 

Second, the controlled problems show that reduction based only on coefficient magnitude, parameter zeroing, or the least-sensitive Fisher direction hinders pruning when simplification requires non-zero limits, asymptotic reductions, parameter identifications, or suppression of strongly identified but removable directions. Reliable structural recovery and pruning efficacy therefore require both a richer catalogue of admissible boundaries and a criterion that goes beyond Fisher sensitivity alone (\cref{fig:cases-summary,tab:cases-comparison}). 

Third, the importance of the acceptance rule depends on the complexity of the models being reduced. Our method and \texttt{bic\_only} selection behave similarly on the controlled suite (\cref{tab:cases-comparison}), where the starting expressions are deliberately redundant but of moderate complexity. Their behaviour separates more clearly on heterogeneous evolved GPSR expressions, where BIC-based pruning yields a less favourable complexity-accuracy trade-off (\cref{tab:pop-pareto_main}) and changes in BIC are not consistently aligned with changes in out-of-sample prediction error. In what follows, we first examine controlled reduction mechanisms and structural recovery, then we turn to population-level pruning of evolved SR models.

\subsection{Controlled diagnostic problems} 
\label{sec:res-controlled} 

\begin{figure}[pos=!htbp]
\centering 
\includegraphics[width=0.98\linewidth]{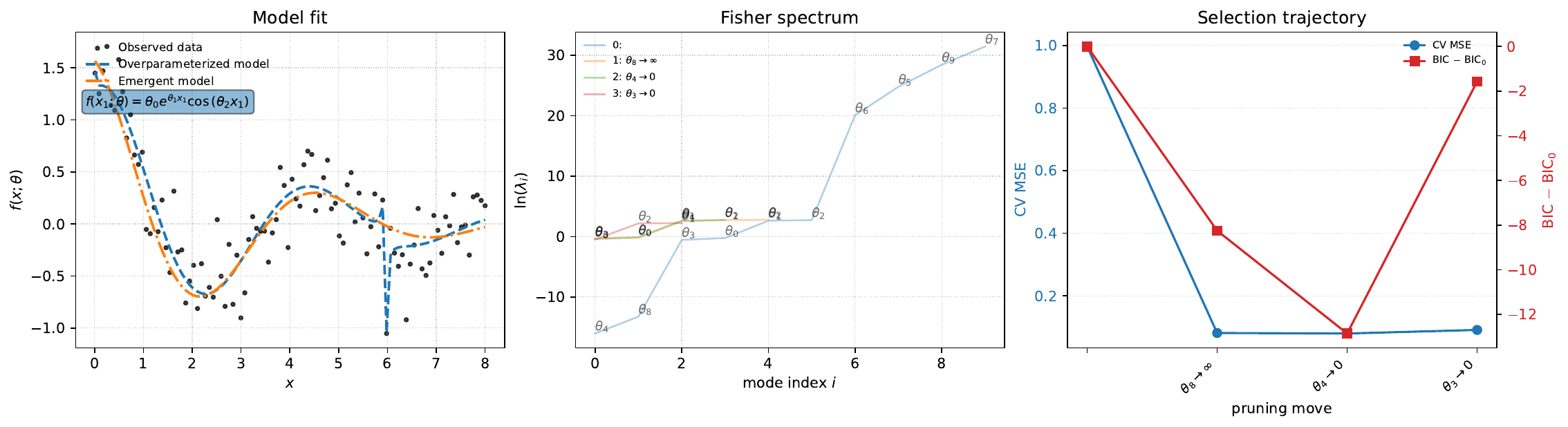}\\[1pt] \includegraphics[width=0.98\linewidth]{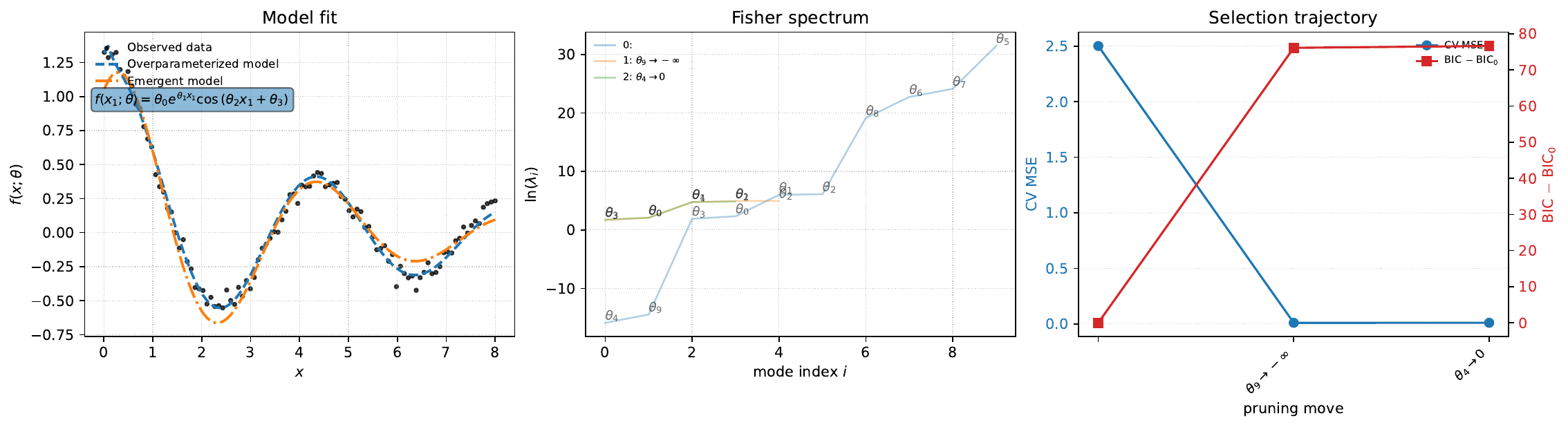} \caption{Problem~\#5, Bessel approximation. Top (bottom) row represents high (low) noise regimes. Left: data and fitted models before and after reduction. Center: ordered $\ln\lambda_i$ of $\Fhat$ after each accepted move. Right: CV and BIC along the reduction trajectory. A single representative run is shown.
The trajectories illustrate how improvements in full-sample fit can fail to persist under fold-wise refitting, leading BIC and cross-validated predictive risk to favour different reductions.} \label{fig:case-bessel} 
\end{figure}

\paragraph{\textbf{An illustrative trajectory.}} 
\Cref{fig:case-bessel} shows a complete reduction trajectory for problem~\#5, i.e., the representative Bessel approximation case study. The panels report the fitted model against the data (left), the evolution of the Fisher spectrum along the reduction path , shown as the ordered $\ln\lambda_i$ after each accepted move (center), and the corresponding CV-BIC selection trajectory (right). 

This plot shows why both finite and infinite reduction constraints are useful: infinite saturation suppresses an entire rational term, whereas zeroing removes standard identifiable additive terms. It also shows that pruning can act at both ends of the Fisher spectrum: the central panel reveals a narrower eigenspectrum for the emergent model, indicating removal of both stiff and sloppy directions. Finally, the example illustrates the role of predictive rather than purely in-sample model selection.  In the high-noise regime, the overparameterised model exploits the rational correction through a near-pole configuration to track individual observations; pruning removes this fragile component and retains the simpler oscillatory-exponential structure. Notably, the same structure is selected at low noise, despite the rational correction strongly improving the full-sample fit: its removal increases BIC by approximately $80$, so a BIC-based rule would strongly favour the unreduced model. Indeed, the unreduced model visibly follows the observations more closely than the emergent model, making the reduction appear counterintuitive. Yet this apparent loss of fit reflects the removal of a flexible component whose contribution is beneficial on the full sample but does not generalise under resampling and refitting. Cross-validation instead confirms this lack of generalisation, as the fitted rational configuration varies across folds and generalises poorly to held-out data (with CV risk decreasing from approximately $2.5$ to $0.02$). The predictive gate therefore favours the simpler, more stable structure, illustrating the rationale of \cref{sec:acceptance}: pruning should preserve generalisable predictive structure rather than sample-specific improvements, especially in the $M$-open regime. Trajectories for the other cases can be found in the supplementary material.

\begin{figure}[pos=!htbp]
\centering 
\includegraphics[width=0.98\linewidth]{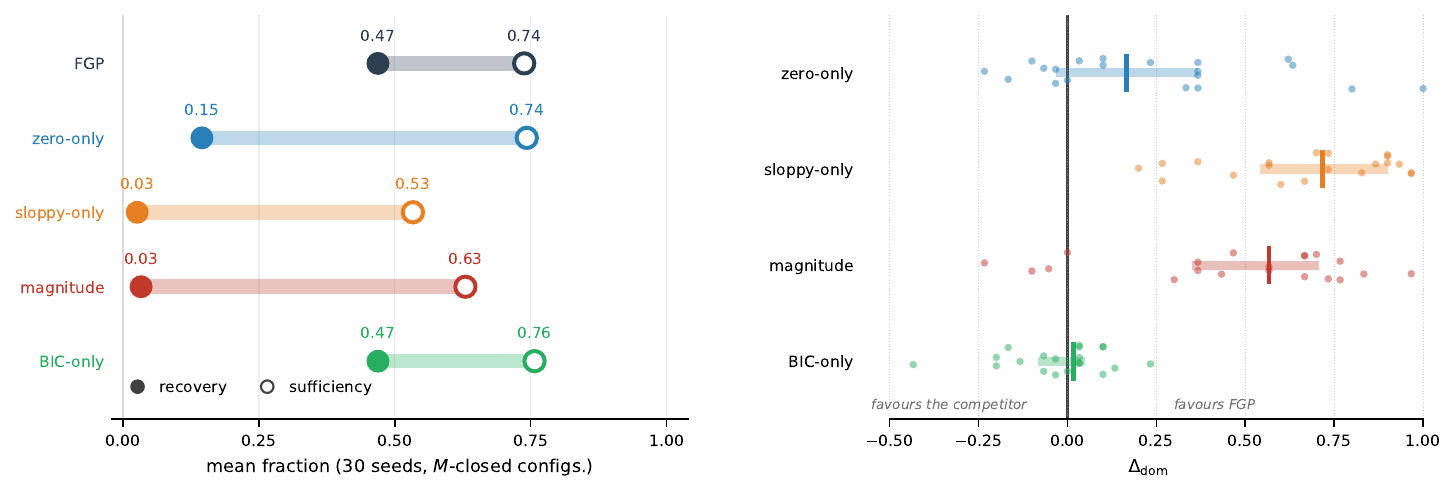} \caption{Aggregate results on the controlled suite. Left: Mean recovery and sufficiency rates over the $14$ $M$-closed configurations. Filled markers denote recovery and open markers sufficiency; their difference measures residual redundancy. Right: net Pareto-dominance rate $\Delta_{\mathrm{dom}}$ on (complexity, $\mathrm{MSE}_{\star}$) over all $20$ configurations. Points denote individual configurations, bars the median, and bands the interquartile range. Positive values favour \texttt{FGP}.}
\label{fig:cases-summary}
\end{figure} 

\begin{table}[!htbp]
\centering
\caption{
Direct paired comparisons of \texttt{FGP} against the competing operators on the controlled suite. Recovery is evaluated over the $14$ $M$-closed configurations; all other metrics use the full set of $20$ configurations. Positive $\Delta_{\mathrm{rec}}$, $r$, and $\Delta_{\mathrm{dom}}$ favour \texttt{FGP}. $p$-values are Holm-corrected within each metric; boldface marks $p_{\mathrm{Holm}}<0.05$.
} 
\label{tab:cases-comparison}
\begin{tabular}{@{}l cc cc cc cc@{}}
\toprule
&
\multicolumn{2}{c}{Recovery}
&
\multicolumn{2}{c}{Complexity}
&
\multicolumn{2}{c}{$\mathrm{MSE}_{\star}$}
&
\multicolumn{2}{c}{Pareto}
\\
\cmidrule(lr){2-3}
\cmidrule(lr){4-5}
\cmidrule(lr){6-7}
\cmidrule(lr){8-9}
Comparison
& median $\Delta_{\mathrm{rec}}$
& $p_{\mathrm{Holm}}$
& $r$
& $p_{\mathrm{Holm}}$
& $r$
& $p_{\mathrm{Holm}}$
& median $\Delta_{\mathrm{dom}}$
& $p_{\mathrm{Holm}}$
\\
\midrule
\texttt{FGP} vs.\ \texttt{zero\_only}
& $+0.33$
& {\boldmath$0.016$}
& $+0.53$
& $0.116$
& $+0.30$
& $0.298$
& $+0.17$
& {\boldmath$0.024$}
\\

\texttt{FGP} vs.\ \texttt{sloppy\_only}
& $+0.40$
& {\boldmath$0.009$}
& $+1.00$
& {\boldmath$4.8\!\cdot\!10^{-4}$}
& $+0.97$
& {\boldmath$3.8\!\cdot\!10^{-5}$}
& $+0.72$
& {\boldmath$3.5\!\cdot\!10^{-4}$}
\\

\texttt{FGP} vs.\ \texttt{magnitude}
& $+0.47$
& {\boldmath$0.009$}
& $+0.75$
& {\boldmath$0.016$}
& $+0.77$
& {\boldmath$0.004$}
& $+0.57$
& {\boldmath$0.001$}
\\

\texttt{FGP} vs.\ \texttt{bic\_only}
& $+0.00$
& $0.873$
& $+0.75$
& $0.116$
& $-0.40$
& $0.298$
& $+0.02$
& $0.519$
\\
\bottomrule
\end{tabular}
\end{table}

\paragraph{\textbf{Behaviour across the suite.}}

We now aggregate results over the complete controlled suite.
\Cref{fig:cases-summary} separates structural recovery from the joint
complexity-accuracy behaviour, while \cref{tab:cases-comparison} reports the corresponding paired comparisons against \texttt{FGP}.

On the $14$ recoverable configurations, \texttt{FGP} and \texttt{bic\_only} attain the same mean recovery rate ($0.47$). The main difference from \texttt{zero\_only} is not sufficiency: both retain a model family containing the truth in $74\%$ of the runs. Rather, \texttt{zero\_only} recovers the parsimonious generating family much less often ($0.15$), indicating that zeroing frequently preserves the relevant structure but cannot complete reductions that require non-zero boundaries. By contrast, \texttt{sloppy\_only} and \texttt{magnitude} also show lower sufficiency, indicating more frequent loss of the generating family itself. Detailed per-problem recovery and sufficiency rates are reported in the supplementary material.

The same pattern appears in the joint complexity-accuracy analysis.
\texttt{FGP} is clearly separated from \texttt{sloppy\_only} and
\texttt{magnitude}, and also has positive net Pareto dominance over
\texttt{zero\_only}. The latter is not significantly separated from
\texttt{FGP} when complexity and $\mathrm{MSE}_{\star}$ are considered individually, but differs under their joint Pareto criterion (\cref{tab:cases-comparison}). Thus, the advantage of the complete operator is primarily associated with the richer candidate-generation and reduction mechanisms rather than with a uniform marginal improvement in either objective.

In contrast, \texttt{bic\_only} remains statistically indistinguishable from \texttt{FGP} across recovery, marginal complexity and prediction error, and joint Pareto dominance. This agreement should be interpreted in light of the controlled design: most problems contain the generating function within the candidate family or its closure, and the starting expressions are deliberately constructed and of moderate complexity rather than heterogeneous products of evolutionary search. The suite therefore contains fewer of the fragile cancellations, near-pole configurations, and sample-specific compensations that can make in-sample penalised criteria diverge from predictive validation. On these controlled models, changing the acceptance rule consequently has little effect once the same candidate reductions are available. The next section tests whether this correspondence persists on complete GPSR populations.

\subsection{Pruning GPSR Pareto fronts}
\label{sec:res-population}

\begin{figure}[pos=!htbp]
\centering 
\includegraphics[width=0.98\linewidth]{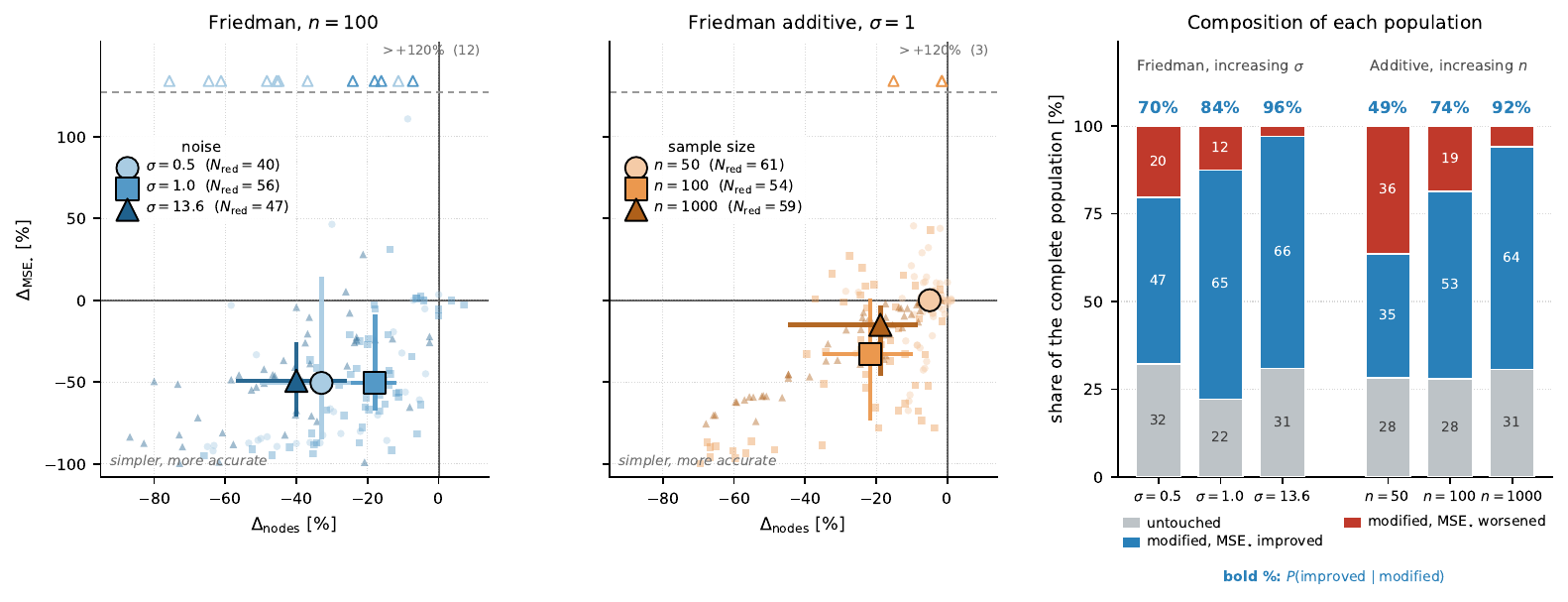}
\caption{Effect of the FGP operator on the six GPSR populations.
Left and center panel: relative percentage change in node count versus noiseless test MSE for expressions on which at least one reduction is accepted. Negative values indicate lower complexity and lower test error, respectively. Small markers denote individual expressions; large markers and bars show the
coordinate-wise medians and interquartile ranges. Open triangles indicate
values clipped for display only. 
Right panel: composition of each complete population into untouched,
modified-improved, and modified-worsened expressions. Labels report
$P(\mathrm{MSE}_{\star}\text{ improves}\mid\text{modified})$.}
\label{fig:population-full}
\end{figure}

\Cref{fig:population-full} summarises the effect of the FGP
operator on the six GPSR populations. Left and central panel show, for every
expression on which at least one reduction is accepted, the relative percentage change in node count against the relative change in noiseless test MSE. The right panel complements this conditional view by showing how frequently the operator acts and, among modified expressions, how often test performance improves.

Overall, pruning improves the complexity--generalisation trade-off rather than
simply reducing model size at the expense of prediction. In five of the six
experimental conditions, the median modified expression moves towards both
lower structural complexity and lower noiseless test error; the only exception
is the smallest-sample regime, where the median predictive effect is essentially
neutral. The detailed behaviour nevertheless differs between the noise and
sample-size series.

For Friedman at fixed $n=100$, the fraction of modified expressions whose
noiseless test MSE improves rises from $70.0\%$ at $\sigma=0.5$ to $83.9\%$
at $\sigma=1$ and $95.7\%$ at $\sigma=13.6$. At the same time, the median
test-MSE reduction remains close to $50\%$ across the three noise levels.
Increasing noise therefore affects primarily how consistently pruning is
beneficial, rather than the typical magnitude of the gain when it succeeds.
This is consistent with noisier evolved expressions containing removable,
sample-specific structure more frequently.

The Friedman additive series shows a complementary dependence on sample size.
At $n=50$, pruning has a comparatively small structural effect and produces
essentially no median change in test MSE, with improved and worsened expressions
almost evenly split ($49.2\%$ improve). At $n=100$ and $n=1000$,
improvements become progressively more frequent ($72.2\%$ and $91.5\%$),
while their median magnitude decreases from $32.8\%$ to $15.0\%$.
Increasing sample size therefore makes accepted reductions more consistently
beneficial, while leaving less predictive excess to remove. This pattern is
not explained simply by the operator acting less often: the untouched fraction
remains broadly comparable across the six populations.

Taken together, the two series indicate that noise and sample size affect
different aspects of pruning efficacy: higher noise is associated with more
frequent gains from removing sample-specific structure, whereas larger samples make accepted reductions more reliably beneficial but typically reduce the magnitude of the attainable improvement.

Finally, we stress that the results above are not confined to a finely tuned hyperparameter configuration. As shown in \Cref{app:res-sensitivity}, although the method depends on three main hyperparameters, $(\beta,\tau,\kappa)$, sensitivity is concentrated in the predictive tolerance $\beta$. This parameter controls the expected trade-off between conservative model retention and more aggressive simplification and should not be set too small, as an overly restrictive predictive gate can prevent useful reductions under sampling fluctuations. The boundary-screen threshold $\tau$ mainly controls the breadth of the candidate set and likewise becomes detrimental only when chosen too restrictively, whereas $\kappa$ has little effect once a non-zero CV-equivalence band is allowed. The same qualitative behaviour is observed on an evolved GPSR population. Overall, the sensitivity analysis identifies mainly lower bounds below which pruning becomes unnecessarily conservative, rather than narrow problem-specific optima. We therefore use the reference configuration $(\beta,\tau,\kappa)=(0.1,16,2)$ unchanged across all experiments, rather than tuning it to individual problems.

\subsubsection{Comparative analysis with other pruning variants}
\label{sec:sr_ablation}

We compare each pruning variant with \texttt{FGP} expression-wise within each GPSR Pareto front. To avoid diluting the comparison with expressions on which no method can act, we remove once per population those left unchanged by all operators; the resulting subset is then fixed across methods. Our primary comparison considers the joint trade-off between final node count and noiseless test MSE through the net Pareto-dominance rate defined in \cref{sec:setup-metrics}.

\begin{table}[!htbp]
\centering
\setlength{\tabcolsep}{4pt}
\caption{Expression-wise Pareto-dominance comparison of each variant against
\texttt{FGP} on the final SR populations, using
$(\mathrm{nodes},\mathrm{MSE}_{\star})$ as objective pairs.
$\Delta_{\mathrm{dom}}$ is the fraction of expressions dominated by
\texttt{FGP} minus the fraction dominated by the variant; parentheses report
the corresponding counts (\texttt{FGP} dominates / variant dominates), with
the remainder being incomparable. Positive values therefore favor
\texttt{FGP}. Boldface indicates statistically significant net dominance in favour of FGP (\(p_{\mathrm{Holm}}<0.05\)).}
\label{tab:pop-pareto_main}
\begin{tabular}{@{}l c rrrr@{}}
\toprule
Population & $N$ & \texttt{zero\_only} & \texttt{sloppy\_only} & \texttt{magnitude} & \texttt{bic\_only}\\
\midrule
Friedman, $\sigma=0.5$ & $42\text{--}43$ & $+0.07\,{\scriptscriptstyle(8/5)}$ & {\boldmath$+0.67\,{\scriptscriptstyle(28/0)}$} & {\boldmath$+0.29\,{\scriptscriptstyle(12/0)}$} & $-0.02\,{\scriptscriptstyle(7/8)}$\\
Friedman, $\sigma=1.0$ & $58$ & $+0.22\,{\scriptscriptstyle(25/12)}$ & {\boldmath$+0.60\,{\scriptscriptstyle(36/1)}$} & $+0.14\,{\scriptscriptstyle(19/11)}$ & $+0.21\,{\scriptscriptstyle(26/14)}$\\
Friedman, $\sigma=13.6$ & $50\text{--}51$ & {\boldmath$+0.27\,{\scriptscriptstyle(23/9)}$} & {\boldmath$+0.34\,{\scriptscriptstyle(27/10)}$} & {\boldmath$+0.54\,{\scriptscriptstyle(28/1)}$} & {\boldmath$+0.33\,{\scriptscriptstyle(25/8)}$}\\
Additive, $n=50$ & $70$ & $-0.16\,{\scriptscriptstyle(9/20)}$ & $+0.17\,{\scriptscriptstyle(24/12)}$ & $-0.20\,{\scriptscriptstyle(13/27)}$ & {\boldmath$+0.23\,{\scriptscriptstyle(25/9)}$}\\
Additive, $n=100$ & $62$ & $+0.16\,{\scriptscriptstyle(21/11)}$ & {\boldmath$+0.23\,{\scriptscriptstyle(18/4)}$} & $-0.08\,{\scriptscriptstyle(12/17)}$ & {\boldmath$+0.35\,{\scriptscriptstyle(30/8)}$}\\
Additive, $n=1000$ & $63\text{--}67$ & {\boldmath$+0.31\,{\scriptscriptstyle(32/11)}$} & {\boldmath$+0.25\,{\scriptscriptstyle(21/4)}$} & {\boldmath$+0.24\,{\scriptscriptstyle(28/12)}$} & {\boldmath$+0.40\,{\scriptscriptstyle(29/4)}$}\\
\bottomrule
\end{tabular}
\end{table}

\Cref{tab:pop-pareto_main} shows that,  across the six populations, the joint complexity-accuracy comparison generally favours \texttt{FGP}. The clearest separation is against \texttt{sloppy\_only}, which is more frequently dominated by \texttt{FGP} across the experimental conditions. \texttt{bic\_only} is more frequently dominated by \texttt{FGP} on the evolved populations, in contrast with the near-equivalence observed on the controlled suite. Moreover, \texttt{zero\_only} remains comparatively competitive in predictive accuracy but tends to retain larger expressions, resulting in a positive net-dominance advantage for \texttt{FGP} in most populations. \texttt{magnitude} exhibits the least stable behaviour: it can occasionally return smaller expressions, but this does not translate into a consistent advantage once complexity and test error are considered jointly.

The corresponding marginal comparisons in node count, number of free parameters, and test MSE, and a further Pareto analysis in the $(|\bth|,\mathrm{MSE}_\star)$ objective space are reported in the supplementary material. They highlight the specific method patterns described above: the separation is strongest against \texttt{sloppy\_only}, \texttt{zero\_only} is mainly disadvantaged in complexity, \texttt{magnitude} exhibits mixed effects, and \texttt{bic\_only} shows clear suboptimal performance also at marginal-level comparisons. 

We would like to stress that the improved complexity-accuracy trade-off of \texttt{FGP} does not come with a comparably large runtime penalty. As detailed in \Cref{app:cost}, wall-clock differences among pruning variants are moderate and \texttt{FGP} is not systematically the most expensive method. Runtime is driven primarily by repeated nonlinear refitting and cross-validation and, empirically, by the number of parameters in the starting expression, with \(t\propto p^{3.6\pm0.2}\) over the investigated GPSR populations. The main practical limitation is therefore scalability with expression dimension rather than a large constant-factor overhead introduced by the richer reduction mechanism.

\section{Conclusions}
\label{sec:conclusion}

This work starts from a simple premise: the complexity of a SR generated
model cannot be diagnosed reliably from its syntax alone. Once its numerical
constants have been fitted, an evolved expression may contain dormant
substructures, weakly identified parameter combinations, saturated nonlinearities,
or redundant parameterisations that contribute little to the effective model.
Conversely, local sensitivity alone does not determine whether structure should
be retained: a locally influential direction may still be compatible with a
simplifying boundary, while a weakly sensitive direction may represent either
removable redundancy or a legitimate subleading contribution. Model reduction
therefore requires distinguishing how strongly the fitted model responds
to a perturbation from whether the fitted solution lies close to a
symbolically meaningful lower-dimensional constraint.

The proposed framework operationalises this distinction by combining local Fisher geometry with explicit boundary-compatibility tests. Geometry is used to diagnose non-identifiable structure and to organise candidate generation, while reductions are expressed directly in the original parameterisation through a small catalogue of symbolically realisable constraints: finite limits, asymptotic limits, and signed parameter identifications. Crucially, none of these local diagnostics certifies a reduction by itself. Every candidate is realised, refitted, and retained only when its cross-validated predictive risk remains within the prescribed tolerance. The resulting operator is therefore best understood as a post-hoc structural regulariser: it does not search for a different symbolic model, but attempts to expose a simpler effective model already contained in an expression returned by SR.

The experiments support both parts of this construction. On the controlled problems, restricting reduction to the sloppiest direction or to coefficient magnitude is markedly less reliable, while zero-boundary pruning is a strong competitor but cannot reproduce reductions that require multiplicative identities, asymptotic limits, or parameter identifications. On complete GPSR populations, the same richer move set remains effective on heterogeneous and generally misspecified expressions: pruning systematically reduces structural complexity while, in most regimes, preserving or improving agreement with the noiseless generating function. The population experiments also reveal a stronger role for predictive validation. Whereas BIC-based and cross-validation-based acceptance often agree on the controlled suite, their behaviour diverges more clearly on evolved expressions, where fragile cancellations, near-singular configurations, and sample-specific improvements are more common. This supports the separation adopted by the method: the boundary catalogue determines which simplifications can be reached, while predictive validation determines which of them remain reliable after refitting. Finally, hyperparameter sensitivity analysis indicates that our pruning method behaviour is not critically dependent on finely tuned values of the geometric screening or tie-breaking parameters, while the predictive acceptance tolerance exposes the expected and interpretable trade-off between aggressiveness of reduction and conservativeness of model retention, and can therefore be used to explicitly control the intended pruning regime.

\paragraph{Limitations and outlook.} The proposed operator remains subject to some limitations. First, reduction is greedy: one move is accepted at a time and the procedure does not backtrack. Consequently, even when the generating function belongs to the starting family or its closure, exact recovery is not guaranteed. Two individually admissible moves may lead to different downstream trajectories, and an early choice can exclude a better final reduction. This is consistent with the objective of the method, which is not to certify recovery of a unique true expression but to obtain a lower-dimensional model while controlling estimated out-of-sample risk. Limited look-ahead or a small beam of competing reduction trajectories would be natural extensions when structural recovery is the primary objective.

Second, the present catalogue of moves is deliberately limited. Zero (and unit) constraints, asymptotic limits, and signed identifications cover several common failure modes of machine-generated expressions, but they are not exhaustive. Other symbolic families may admit useful constraints requiring different reparameterisations or substitutions, while reductions involving arbitrary linear combinations of parameters remain outside the current implementation. The general framework is nevertheless extensible: additional constraints can be incorporated whenever they admit a meaningful compatibility statistic and an explicit symbolic realisation.

A crucial limitation is the dependence on reliable numerical optimisation. The Fisher matrix, identifiability diagnosis, and boundary statistics are all evaluated at the fitted solution, so optimisation failures can propagate into the reduction trajectory. We deliberately adopted conservative multi-start and warm-started nonlinear fitting to minimise this source of error, at the cost of substantial computation. In its present form, the operator is therefore better suited to post-hoc reduction of a restricted set of final expressions than to routine application to every individual at every SR generation. Lighter optimisation strategies could make selective use during evolution feasible when occasional fitting failures can be tolerated. Moreover, our method could be combined with semantics-preserving algebraic simplification \citep{defranca2025egraph}, a complementary approach to be applied before pruning to reduce the dimension of the geometric problem or afterwards to expose the reduced model in a cleaner syntactic form and prevent spurious numerical instabilities.

Overall, the results suggest that complexity control in SR need not end with the objective used during evolutionary search. An evolved expression can itself be treated as an overparameterised fitted model and interrogated for the lower-dimensional structure supported by the data. By combining information geometry, interpretable parametric boundaries, symbolic realisation, and predictive validation, the proposed framework turns this idea into an automatic reduction operator that improves the complexity-accuracy trade-off across both controlled models and complete SR populations. More broadly, it provides a bridge between information-geometric model reduction and symbolic learning: rather than equating interpretability with the expression originally returned by the search, it asks which part of that expression remains necessary after fitting and predictive validation.\\

\noindent\textbf{\emph{Declaration of generative AI and AI-assisted technologies in the manuscript preparation process.}} During the preparation of this work, the authors used OpenAI ChatGPT to assist with manuscript organisation and language editing, and Anthropic Claude to assist with code generation and implementation. All AI-assisted output was reviewed, edited, and tested by the authors. The authors approved the final manuscript and take full responsibility for its content and for the correctness of the reported analyses.\\

\noindent\textbf{\emph{Funding.}} This research did not receive any specific grant from funding agencies in the public, commercial, or not-for-profit sectors.

\printcredits

\bibliographystyle{cas-model2-names}

\bibliography{emergent}

\appendix
\counterwithin{figure}{section}
\counterwithin{table}{section}
\counterwithin{equation}{section}

\section{Derivations}
\label{app:derivations}

\subsection{Closest point on a linearised constraint}

Minimising the squared Fisher displacement length $\bdelta^{\top}\Fhat_{[n]}\,\bdelta$
subject to the linearised constraint
$c(\hth)+\nabla c(\hth)^{\top}\bdelta=0$ leads to the Lagrangian
\[
L(\bdelta,\mu)=\bdelta^{\top}\Fhat_{[n]}\,\bdelta
+\mu\big(c(\hth)+\nabla c(\hth)^{\top}\bdelta\big),
\]
where $\mu$ is the multiplier.
Stationarity in $\bdelta$ yields $2\Fhat_{[n]}\bdelta+\mu\nabla c(\hth)=0$,
hence $\bdelta=-\tfrac{\mu}{2}\Fhat_{[n]}^{-1}\nabla c(\hth)$.
Substituting into the constraint and solving for $\mu$ gives
\[
c(\hth)-\tfrac{\mu}{2}\nabla c(\hth)^{\top}\Fhat_{[n]}^{-1}\nabla c(\hth)=0
\quad\Longrightarrow\quad
\mu=\frac{2\,c(\hth)}{\nabla c(\hth)^{\top}\Fhat_{[n]}^{-1}\nabla c(\hth)} ,
\]
so that the minimising displacement and its squared length are
\[
\delta^{\star}=-\frac{c(\hth)}{\nabla c(\hth)^{\top}\Fhat_{[n]}^{-1}\nabla c(\hth)}\,
\Fhat_{[n]}^{-1}\nabla c(\hth)
\quad\mbox{and}\quad
(\delta^{\star})^{\top}\Fhat_{[n]}\,\delta^{\star}
=\frac{c(\hth)^{2}}{\nabla c(\hth)^{\top}\Fhat_{[n]}^{-1}\nabla c(\hth)},
\]
which results in \cref{eq:fisherdistance}.

\subsection{Wald calibration}

Under standard regularity conditions $\hth\approx\mathcal{N}(\bth,\Fhat_n^{-1})$.
Then, a first-order expansion of $c$ around $\bth$ results in
$c(\hth)\approx c(\bth)+\nabla c(\bth)^{\top}(\hth-\bth)$,
and the delta method yields
\[
c(\hth)\;\approx\;
\mathcal{N}\!\Big(c(\bth),\;
\nabla c(\bth)^{\top}\Fhat_n^{-1}\nabla c(\bth)\Big).
\]
Replacing $\bth$ by $\hth$ gives the plug-in variance estimator
$\widehat{\mathrm{Var}}[c(\hth)]=\nabla c(\hth)^{\top}\Fhat_n^{-1}\nabla c(\hth)$,
so that, under $H_0:c(\bth)=0$, the statistic
$Z=c(\hth)/\widehat{\mathrm{Var}}[c(\hth)]^{1/2}$ is approximately standard normal
and $Z^{2}=\dF$ is approximately $\chi^{2}_{1}$
\citep{wald1943tests}.

\subsection{Pullback of the metric under the arc tangent compactification}

Let $\phi_k=\arctan\theta_k$ and $\phi_h=\theta_h$ for $h\ne k$.
The Jacobian $J=\partial\bth/\partial\bphi$ is diagonal with $J_{kk}=1+\tan^{2}\phi_k=1+\theta_k^{2}$
and $J_{hh}=1$ for $h\ne k$.
The metric transforms as $\Fis_{\bphi}=J^{\top}\Fis_{\bth}\,J$ and,
since $J$ is diagonal and invertible, this results in
$\Fis_{\bphi}^{-1}=J^{-1}\Fis_{\bth}^{-1}J^{-1}$,
whence
\[
\big(\Fis_{\bphi}^{-1}\big)_{kk}
=\frac{(\Fis_{\bth}^{-1})_{kk}}{(1+\theta_k^{2})^{2}},
\]
while $\big(\Fis_{\bphi}^{-1}\big)_{hh}=\big(\Fis_{\bth}^{-1}\big)_{hh}$ for $h\neq k$.
With $c(\phi)=\phi_k\mp\frac{\pi}{2}$ and $\nabla c=e_k$ in $\phi$-coordinates,
\cref{eq:fisherdistance} gives
\[
\dF
=\frac{\big(\arctan\hat\theta_k\mp\frac{\pi}{2}\big)^{2}\big(1+\hat\theta_k^{2}\big)^{2}}
      {\big(\Fhat^{-1}_{[n]}\big)_{kk}},
\]
which is the entry in \cref{tab:boundaries}.
Note that the pullback is exact for the metric, but the constraint linearisation is exact only because $c$ is linear
in $\phi$; what remains approximate is the constancy of the metric along the (now finite) displacement.

\section{Algorithm details}

\subsection{Subroutine specifics}
\label{app:routines}
Further subroutines used in the \cref{alg:main} are defined in \cref{alg:screen}, \cref{alg:reduce}, and \cref{alg:select}.

\begin{algorithm}[t]
\caption{\textsc{Screen} - candidate parameters and their boundary menus for mode $m$}
\label{alg:screen}
\begin{algorithmic}[1]
\Require $\mathcal{G}$, mode $m$, thresholds $\epsilon_\lambda,\Delta_\lambda,\tau$
\Ensure candidate set $\mathcal{K}=\{(\theta_k,\mathfrak{B}_k)\}$
\State $L\gets\mathcal{U}$ \Comment{nominated by \cref{eq:lowmask,eq:participation}}
\State $H\gets\{1,\dots,p\}\setminus L$ \Comment{one pool for every mode}
\If{$m=\textsc{unidentifiable}$}
   \State $\mathcal{K}\gets\{(\theta_k,\{0,1,+\infty,-\infty\})\mid k\in L\}$
   \State augment with $\{(\theta_k,\{s\theta_h\})\mid (h,k,s)\in\mathcal{P}, \{h,k\}\cap L\neq\varnothing \}$
\ElsIf{$m=\textsc{zeroing}$}
   \State $\mathcal{K}\gets\{(\theta_k,\{0\})\mid k\in H,\ d^{2}_{0}(k)<\tau\}$
\ElsIf{$m=\textsc{saturating}$}
   \State $\mathcal{K}\gets\{(\theta_k,\{\sgn(\hat\theta_k)\infty\})\mid k\in H,\
          d^{2}_{\sgn(\hat\theta_k)\infty}(k)<\tau\}$
\Else \Comment{$m=\textsc{symmetric}$}
  \State $\mathcal{K}\gets\{(\theta_k,\{s\theta_h\})\mid (h,k,s)\in\mathcal{P},\
          d^{2}_{\mathrm{sgn}}<\tau,\ h\in H\}$
\EndIf
\State \Return $\mathcal{K}$
\end{algorithmic}
\end{algorithm}

\begin{algorithm}[t]
\caption{\textsc{BestReduction} - evaluate the boundary menu of one parameter}
\label{alg:reduce}
\begin{algorithmic}[1]
\Require candidate $(\theta_k,\mathfrak{B}_k)$, data $\D$, current $f,\hth,c,B$
\Ensure representative move for $\theta_k$
\State $\mathcal{R}\gets\varnothing$
\For{$b\in\mathfrak{B}_k$}
   \State \bf~$g\gets\textsc{Contract}(f|\theta_k=b)$
          \Comment{\cref{sec:method-contraction}}
   \State \bf~$\hth_g\gets\textsc{Fit}(g,\D)$ \Comment{loose screening tolerances}
   \State \bf~$(c_g,s_g)\gets\textsc{CvRisk}(g,\hth_g,\D)$;\quad $B_g\gets\textsc{Bic}(g,\hth_g,\D)$
   \State \bf~$\mathcal{R}\gets\mathcal{R}\cup\{(b,\ \Delta c=c_g-c,\ s_g,\ \Delta B=B_g-B,\ g,\hth_g)\}$
\EndFor
\State \Return $\textsc{Select}(\mathcal{R},c)$ \Comment{same rule, ranked-best move returned}
\end{algorithmic}
\end{algorithm}

\begin{algorithm}[t]
\caption{\textsc{Select} - CV admissibility, then BIC tie-break}
\label{alg:select}
\begin{algorithmic}[1]
\Require moves $\mathcal{R}=\{(b_m,\Delta c_m,s_m,\Delta B_m,\dots)\}$, current CV risk $c$
\Ensure chosen move, accept flag
\State $\mathcal{A}\gets\{m:\Delta c_m \text{ finite},\ \Delta c_m\le\beta\,c\}$
       \Comment{\cref{eq:cv-accept}}
\If{$\mathcal{A}=\varnothing$}\ \Return $(\varnothing,\textsc{false})$\EndIf
\State $m^{\star}\gets\argmin_{m\in\mathcal{A}}\Delta c_m$
\State $\mathcal{E}\gets\{m\in\mathcal{A}:\Delta c_m-\Delta c_{m^{\star}}\le\kappa\,s_{m^{\star}}\}$
       \Comment{\cref{eq:equivset}}
\State \Return $\big(\argmin_{m\in\mathcal{E}}\Delta B_m,\ \textsc{true}\big)$
       \Comment{ties by move priority, then lexicographic}
\end{algorithmic}
\end{algorithm}

\subsection{Numerical realisation of the fits}
\label{app:numerics}
\paragraph{Full-data fitting.}
Every diagnostic fit is obtained with a bound-constrained trust-region-reflective least-squares solver \citep{branch1999subspace,virtanen2020scipy}, applied to the penalised residual
stack
\[
\big[\,y-f(x;\bth)\,;\ \sqrt{\varrho}\,\bth\,\big].
\]
Optimisation follows a decreasing ridge-continuation path $\varrho\in\{10^{-2},10^{-4}\}$, followed by an almost unpenalised polish at $\varrho=10^{-8}$. The polish is retained only if it does not increase the unpenalised residual sum of squares and satisfies a conditioning guard,
allowing $\mathrm{cond}(J^\top J)$ to deteriorate by at most six orders of magnitude.

Starts consist of a warm start (when available), the midpoint of the parameter box, the origin, and $n_{\mathrm{starts}}=30$ independent Gaussian initialisations drawn from $\mathcal{N}(0,0.1^2)$. Parameter bounds are $[-10^2,10^2]$ for the controlled diagnostic problems and $[-10^8,10^8]$ for the GPSR Pareto-front experiments.

\paragraph{Screening and diagnostic fits.}
Candidate screening uses loose tolerances ($10^{-4}$, at most $500$ function
evaluations, without the final polish). Each screening fit is attempted twice:
once from the parent model's fitted solution and once from a fresh random
initialisation, retaining the solution with lower training RMSE. The second
attempt reduces the risk that a warm start remains trapped near a degenerate
configuration after an asymptotic reduction.

Once a move is accepted, the resulting model is refitted using the full
multi-start procedure with tolerances $10^{-10}$ and at most $5000$ function
evaluations before its Fisher spectrum and boundary statistics are recomputed.
Hence, the geometry used to generate the next reduction is never computed from
a screening-quality fit. Cross-validation fits use the screening tolerances and
are warm-started from the candidate's full-data solution, which substantially
reduces the cost of the $K$ fold-specific refits.

\section{Controlled diagnostic problems definition}
\label{app:cases}

This section gives the complete specification of the ten controlled diagnostic problems used in the first experimental study.  Each problem is defined by a truth $f^{\star}$, a candidate expression $\hat f(\bth,x)$ that either contains $f^{\star}$ as a special case ($M$-closed) or can only approximate it ($M$-open), a sampling domain, and two noise levels. The training sample size is $n=100$ throughout, and the observational noise is additive Gaussian, $\varepsilon\sim\mathcal{N}(0,\sigma^{2})$.
All definitions per case study are reported in \Cref{tab:cases-def}, including the sampling domain, the number of free parameters, the two noise levels, and the corresponding $M$-regime. The problems were selected to separate three aspects of the reduction task: the statistical regime of the fitted model, the symbolic move required to obtain the intended reduction, and additional numerical pathologies that may interfere with the geometric diagnosis.

\begin{table}[!htbp]
\centering\footnotesize
\setlength{\tabcolsep}{2.5pt}
\caption{Definitions of the ten controlled diagnostic problems. $p$ denotes
the number of free parameters in the candidate expression; $n=100$ in all
cases. The columns $\sigma_{\mathrm{lo}}/\sigma_{\mathrm{hi}}$ report the two
observational noise levels. Inputs are sampled on a regular grid.
The $M$-regime column uses closed$^\star$ to indicate that the
data-generating function belongs only to the closure of the candidate family and
is recovered through an asymptotic parameter limit.}
\label{tab:cases-def}
\renewcommand{\arraystretch}{2.0}
\setlength{\tabcolsep}{2.5pt}
\resizebox{\linewidth}{!}{%
\begin{tabular}{@{}c l p{2.15cm} p{5.5cm} p{1.7cm} c c c@{}}
\toprule
\# & Name & Truth $f^{\star}(x)$ & Candidate $\hat f(\bth,x)$ & Domain & $p$ &
$\sigma_{\mathrm{lo}}/\sigma_{\mathrm{hi}}$ & $M$-reg.\\
\midrule
1 & Nested sine--quadratic
& $0.7x^{2}+2\sin(1.5x)$
& $\theta_0\sin(\theta_1x)+\theta_2x^{2}+\theta_3x^{3}
   +\dfrac{\theta_4}{1+e^{-\theta_5x}}+e^{\theta_6x+\theta_7}$
& $x\in(0,5]$ & 8 & $0.5/1.5$ & closed$^\star$ \\
2 & Expansion 1D
& $e^{0.7x}$
& $\displaystyle\sum_{h=0}^{5}\theta_hx^{h}+\theta_6\sin(\theta_7x)
   +\dfrac{\theta_8}{1+e^{-\theta_9x}}$
& $x\in(0,1.5]$ & 10 & $0.05/0.2$ & open \\
3 & Expansion 2D
& $e^{0.7x_1}$
& $ \displaystyle\sum_{h\in\{0,1,2,3,5\}}\theta_hx_1^{h}+\theta_4x_2^{4}
   +\theta_6\sin(\theta_7x_1)+\dfrac{\theta_8}{1+e^{-\theta_9x_2}}$
& $x_1\!\in\!(0,1.5]$, $x_2\!\in\!(0,2.5]$ & 10 & $0.05/0.2$& open  \\
4 & Near-alias Fourier
& $\sin(0.65x+0.5)$
& $\theta_0\sin(\theta_1x+\theta_2)+\theta_3\sin(\theta_4x)+\theta_5+\theta_6x$
& $x\in[0,30]$ & 7 & $0.05/0.2$& closed  \\
5 & Bessel approximation
& $1.3\,J_0(1.6x)$
& $\theta_0e^{\theta_1x}\cos(\theta_2x+\theta_3)+\theta_4
   +\dfrac{\theta_5+\theta_6x+\theta_7x^{2}}{\theta_8x+\theta_9x^{2}}$
& $x\in[0.01,8]$ & 10 & $0.05/0.3$& open  \\
6 & Rational function
& $\dfrac{1.5x}{1+0.8x}$
& $\dfrac{\theta_0x+\theta_1+\theta_2x^{2}}{\theta_3+\theta_4x+\theta_5x^{2}}
   +\dfrac{\theta_6x}{\theta_7+\theta_8x}$
& $x\in[0,3]$ & 9 & $0.01/0.2$& closed \\
7 & Competing saturations
& $\dfrac{1}{1+e^{-1.2x+1.5}}$
& $\theta_0+\dfrac{\theta_1+\theta_2\log(\theta_3+e^{\theta_4x})}
   {\theta_5+\theta_6e^{\theta_7x}}$
& $x\in[-6,6]$ & 8 & $0.01/0.1$& closed  \\
8 & Separable 2D
& $\dfrac{e^{0.6x_1}}{0.66+1.5\,e^{0.9x_2}}$
& $\theta_0e^{\theta_1x_1}\Big(\theta_2+\theta_3x_1+\theta_4\tfrac{x_1}{x_2}
   +e^{\theta_5x_2+\theta_6x_1+\theta_7}\Big)^{-1}+\theta_8+\theta_9x_1$
& $x_1,x_2\!\in\!(0,3]$ & 10 & $0.01/0.1$& closed \\
9 & Buried transcendentals
& $0.5x+0.7x^{2}$
& $\theta_0x+\theta_1x^{2}+e^{\theta_2x}+\dfrac{1}{1+e^{\theta_3x}}
   +\theta_4x^{3}+\theta_5x^{4}+\theta_6\sin x$
& $x\in[0.05,5]$ & 7 & $0.20/1.5$& closed$^\star$ \\
10 & Tanh gauge
& $1.2\tanh(2x)$
& $\theta_0\dfrac{e^{\theta_1x}-e^{\theta_2x}}{e^{\theta_3x}+e^{\theta_4x}}
   +\theta_5+\theta_6x+\theta_7x^{2}+\theta_8\cos x$
& $x\in[-3,3]$ & 9 & $0.05/0.15$& closed \\
\bottomrule
\end{tabular}}
\end{table}

From a statistical perspective, Problem~\#1 has regular Fisher geometry; Problems~\#2-\#4 and~\#9 are near-singular because of irrelevant directions or approximate parameter symmetries; and Problems~\#5--\#8 and~\#10 contain exact non-identifiabilities and hence singular Fisher matrices. Problems~\#2, \#3, and~\#5 are additionally $M$-open, so their intended reductions are effective approximations rather than exact recovery of a submodel. Problems marked closed$^\star$ contain the truth only in the closure of the candidate family and therefore require an asymptotic parameter limit.

The suite also exercises the four classes of symbolic moves used by the reduction operator. Zero limits remove dispensable additive or multiplicative components and occur in several problems, including \#1-\#3. Unit constraints are required in the scale-redundant rational structures of Problems~\#6-\#8, where common numerator-denominator rescalings generate exact gauges. Asymptotic limits are necessary when an unwanted component has no multiplicative coefficient that can be set directly to zero, as in the exponential contribution of Problem~\#1 and, more prominently, the buried transcendental terms of Problem~\#9. Problem~\#10 requires signed parameter identification to resolve the gauge in the exponential ratio and recover the equivalent hyperbolic-tangent representation. Problems~\#5, \#6, and~\#8 additionally admit pole configurations inside or close to the sampling domain, providing a separate test of whether genuine boundary reductions can be distinguished from numerically unstable rational trajectories.

Examples of the reduction paths for all controlled problems are reported in the Supplementary material.

\section{Sensitivity to hyperparameters}
\label{app:res-sensitivity}

We assess sensitivity to the three main decision parameters of
\texttt{FGP}: the CV acceptance tolerance $\beta$, the boundary-screen
threshold $\tau$, and the CV-equivalence band $\kappa$. Using the reference
configuration of \cref{sec:setup-hyper},
$(\beta,\tau,\kappa)=(0.1,16,2)$, we vary one parameter at a time over
$\beta\in\{0,0.02,0.05,0.1,0.2,0.5\}$,
$\tau\in\{4,9,16,49\}$,
$\kappa\in\{0,1,2,3\}$. Each value is evaluated over the ten controlled problems, two noise levels,
and $30$ seeds, yielding $600$ paired units, with all remaining parameters
held fixed.

\begin{figure}[pos=!htbp]
\centering
\includegraphics[width=\linewidth]{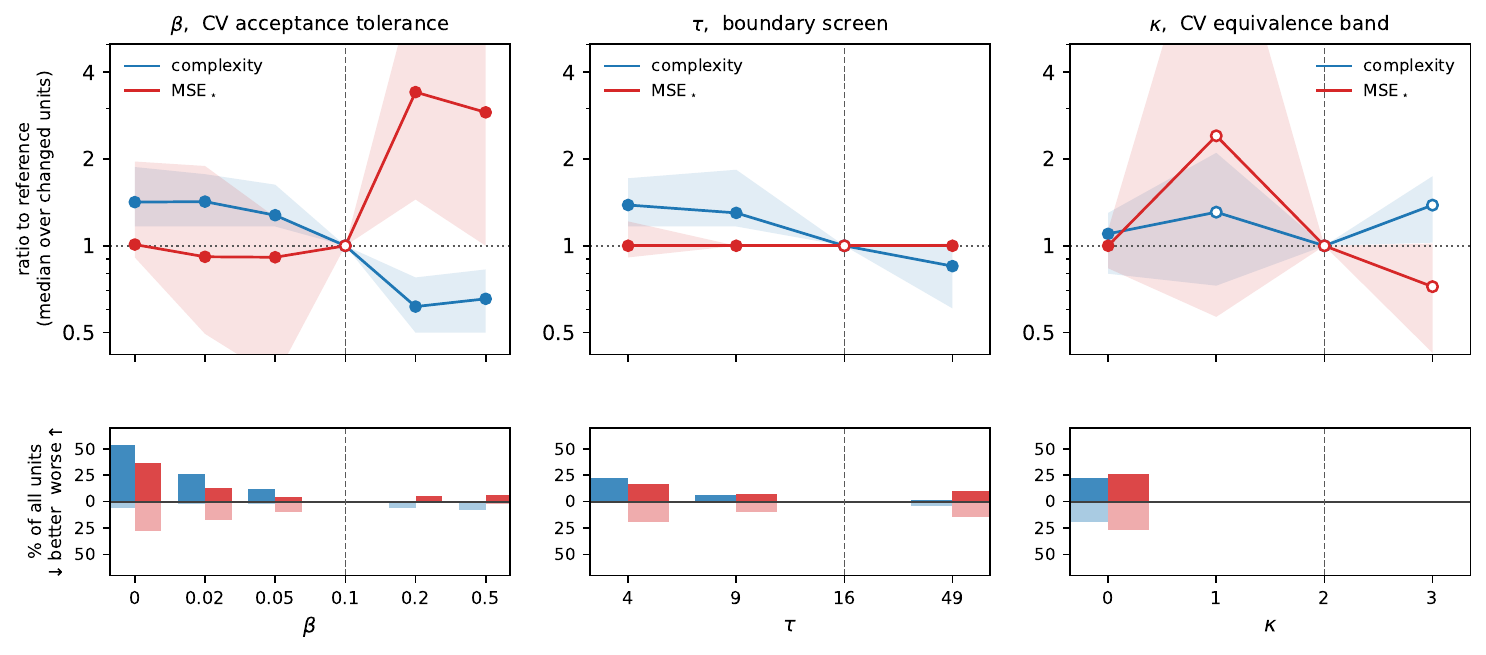}
\caption{Sensitivity of \texttt{FGP} on the controlled problems.
Top: median ratio to the reference configuration over units whose outcome
changes; bands show the interquartile range and hollow markers denote settings
affecting fewer than $5\%$ of all units. Bottom: fractions of all units for
which each metric increases or decreases. The two rows therefore show effect
magnitude and frequency, respectively.}
\label{fig:cases-robust}
\end{figure}

Because many hyperparameter changes return exactly the same expression,
\cref{fig:cases-robust} reports both the magnitude and the frequency of their
effect: the upper panels show the median ratio to the reference configuration
over changed units only, while the lower panels show the fractions of all
units for which complexity or $\mathrm{MSE}_{\star}$ increases or decreases.

Sensitivity is concentrated in $\beta$. Tightening the predictive gate below
the reference value changes more than $50\%$ of the units, generally retaining
more complex models and increasing test error, whereas relaxing it beyond
$\beta=0.1$ affects only about $5\%$. In the relatively small subset where
additional reductions become admissible, complexity can fall to roughly one
quarter of its reference value. The effect of $\tau$ is weaker and primarily
structural: small values restrict the candidate set and therefore lead to less
aggressive pruning, while test performance remains essentially unchanged.
Finally, $\kappa$ behaves almost as a binary switch: $\kappa=0$ removes the BIC
tie-break and yields slightly more complex solutions, whereas
$\kappa\geq1$ produces very similar outcomes across the tested values.

To verify that the same behaviour extends to evolved SR models, we repeat the
three sweeps on the $72$ expressions of the Friedman $\sigma=1.0$ final Pareto
front. Each setting is compared expression-wise with the same reference
configuration, using the protocol of \cref{sec:sr_ablation}; expressions left
unchanged by both settings are excluded from the corresponding comparison.
\Cref{tab:pop-robust} reports the results.

\begin{table}[!htbp]
\centering\footnotesize
\setlength{\tabcolsep}{2pt}
\caption{Sensitivity analysis of the \texttt{FGP} operator on the Friedman \(\sigma=1.0\) SR Pareto front. Each hyperparameter setting is compared expression-wise with the reference configuration. Entries report the median paired difference \(m_{\mathrm{grid}}-m_{\mathrm{ref}}\), computed over expressions for which the two configurations yield different outcomes. Expressions left unreduced by both configurations are excluded from the corresponding comparison.}
\label{tab:pop-robust}
\resizebox{\linewidth}{!}{%
\begin{tabular}{@{}lrrrr@{}}
\toprule
Configuration & $n$ & median $\Delta$ ($r$) & worse/tie/better & $p_{\text{Holm}}$ \\
\midrule
\multicolumn{5}{@{}l}{\textit{complexity -- node count}}\\
\multicolumn{5}{@{}l}{\quad $\beta$: CV acceptance tolerance (reference $\beta=0.1$)}\\
\qquad $\beta=0$ & $51$ & $+6$ $(+0.88)$ & $36/13/2$ & $\mathbf{1.2\!\cdot\!10^{-5}}$\\
\qquad $\beta=0.02$ & $53$ & $+4$ $(+0.78)$ & $32/18/3$ & $\mathbf{2.0\!\cdot\!10^{-4}}$\\
\qquad $\beta=0.05$ & $53$ & $+4$ $(+0.58)$ & $16/32/5$ & $\mathbf{0.019}$\\
\qquad $\beta=0.2$ & $52$ & $-10$ $(-0.83)$ & $3/34/15$ & $\mathbf{0.004}$\\
\qquad $\beta=0.5$ & $52$ & $-17$ $(-0.86)$ & $4/23/25$ & $\mathbf{2.0\!\cdot\!10^{-4}}$\\
\addlinespace[2pt]
\multicolumn{5}{@{}l}{\quad $\tau$: boundary screen (reference $\tau=16$)}\\
\qquad $\tau=4$ & $53$ & $+5$ $(+0.65)$ & $16/34/3$ & $\mathbf{0.026}$\\
\qquad $\tau=9$ & $52$ & $-1$ $(-0.36)$ & $4/43/5$ & $0.343$\\
\qquad $\tau=49$ & $54$ & $-8$ $(-0.76)$ & $3/35/16$ & $\mathbf{0.011}$\\
\addlinespace[2pt]
\multicolumn{5}{@{}l}{\quad $\kappa$: CV equivalence band (reference $\kappa=2$)}\\
\qquad $\kappa=0$ & $53$ & $+2$ $(+0.15)$ & $16/28/9$ & $1.000$\\
\qquad $\kappa=1$ & $54$ & $+2$ $(+0.22)$ & $7/42/5$ & $1.000$\\
\qquad $\kappa=3$ & $52$ & $-11$ $(-0.53)$ & $2/43/7$ & $0.464$\\
\addlinespace[2pt]
\bottomrule
\end{tabular}\hfill
\begin{tabular}{@{}lrrrr@{}}
\toprule
Configuration & $n$ & median $\Delta$ ($r$) & worse/tie/better & $p_{\text{Holm}}$ \\
\midrule
\multicolumn{5}{@{}l}{\textit{$\mathrm{MSE}_{\star}$ -- held-out test MSE}}\\
\multicolumn{5}{@{}l}{\quad $\beta$: CV acceptance tolerance (reference $\beta=0.1$)}\\
\qquad $\beta=0$ & $51$ & $+0.0957$ $(+0.51)$ & $29/11/11$ & $\mathbf{0.020}$\\
\qquad $\beta=0.02$ & $53$ & $+0.139$ $(+0.62)$ & $27/17/9$ & $\mathbf{0.006}$\\
\qquad $\beta=0.05$ & $53$ & $+0.128$ $(+0.57)$ & $17/30/6$ & $0.053$\\
\qquad $\beta=0.2$ & $52$ & $-0.0831$ $(-0.25)$ & $7/33/12$ & $0.490$\\
\qquad $\beta=0.5$ & $52$ & $+0.0928$ $(+0.24)$ & $19/22/11$ & $0.490$\\
\addlinespace[2pt]
\multicolumn{5}{@{}l}{\quad $\tau$: boundary screen (reference $\tau=16$)}\\
\qquad $\tau=4$ & $53$ & $+0.00524$ $(+0.31)$ & $16/26/11$ & $0.225$\\
\qquad $\tau=9$ & $52$ & $\approx\!0$ $(-0.49)$ & $4/34/14$ & $0.212$\\
\qquad $\tau=49$ & $54$ & $\approx\!0$ $(-0.36)$ & $11/28/15$ & $0.225$\\
\addlinespace[2pt]
\multicolumn{5}{@{}l}{\quad $\kappa$: CV equivalence band (reference $\kappa=2$)}\\
\qquad $\kappa=0$ & $53$ & $-0.0461$ $(-0.13)$ & $12/27/14$ & $1.000$\\
\qquad $\kappa=1$ & $54$ & $+0.0017$ $(+0.00)$ & $6/42/6$ & $1.000$\\
\qquad $\kappa=3$ & $52$ & $-0.168$ $(-0.33)$ & $3/43/6$ & $1.000$\\
\addlinespace[2pt]
\bottomrule
\end{tabular}}

\end{table}

The GPSR population reproduces the controlled-suite pattern. Sensitivity is
again dominated by $\beta$: overly restrictive values inhibit useful
simplification, whereas relaxing the gate beyond the reference value reduces
complexity without a systematic deterioration in test MSE. The effects of
$\tau$ and $\kappa$ are substantially weaker and mainly confined to
complexity. Overall, the analysis identifies restrictive lower ranges for
$\beta$ and $\tau$, rather than narrow problem-specific optima, while $\kappa$
has little influence once a non-zero equivalence band is allowed.

\section{Measured cost and scaling}
\label{app:cost}

\begin{figure}[pos=!htbp]
\centering
\includegraphics[width=\linewidth]{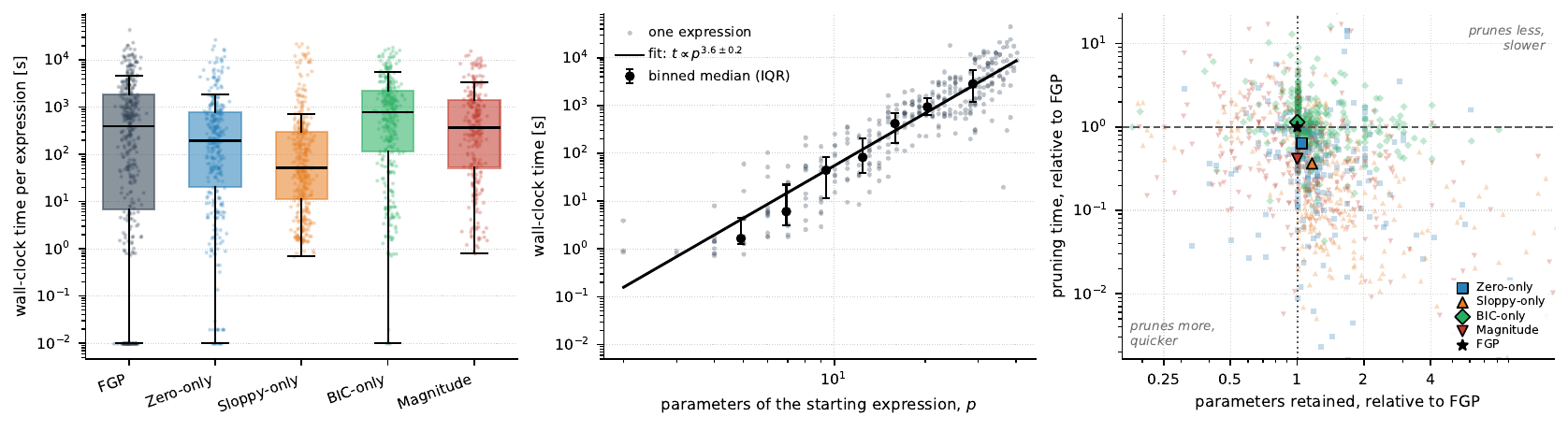}
\caption{Computational cost on the six Operon Pareto front. Left: wall-clock pruning time per expression across methods. Blocks show medians and interquartile ranges.  Center: runtime of \texttt{FGP} versus the number $p$ of parameters in the starting expression; black markers show binned medians and interquartile ranges. Right: expression-wise retained parameter count and runtime of each variant relative to \texttt{FGP}; large markers show per-variant medians.}
\label{fig:cost}
\end{figure}

We examine the computational cost of the pruning operators. On the controlled suite ($10$ problems $\times$ $2$ noise levels $\times$ $30$ runs), the median wall-clock time of \texttt{FGP} is $17$\,s per run (IQR $11$-$31$\,s). The corresponding values are $21$\,s ($15$-$33$\,s; $1.27\times$ \texttt{FGP}) for \texttt{zero\_only}, $10$\,s ($6$-$23$\,s; $0.62\times$) for \texttt{sloppy\_only}, $19$\,s ($14$--$39$\,s; $1.15\times$) for \texttt{bic\_only}, and $13$\,s ($3$-$29$\,s; $0.79\times$) for \texttt{magnitude}. Thus, runtime differences among methods are moderate and \texttt{FGP} is not the most expensive operator. These measurements are single-threaded per task, with candidate evaluations not parallelised within a run. \Cref{fig:cost} shows the corresponding behaviour on the six GPSR populations. The left panel shows that the overall runtime distributions remain broadly comparable across operators. The right panel complements this with an
expression-wise comparison of computational cost and retained parameter count relative to FGP: \texttt{bic\_only} tends to be the most expensive variant without a corresponding systematic gain in parsimony, whereas \texttt{zero\_only} and \texttt{sloppy\_only} are generally cheaper but retain more parameters; \texttt{magnitude} again shows less systematic behaviour relative to \texttt{FGP}. The central panel in \Cref{fig:cost} shows that the dominant practical scaling factor in our reduction method is the number $p$ of parameters in the starting expression. A log-log regression over the pooled GPSR populations gives $t \propto p^{\,3.6\pm0.2}$.
This empirical exponent is compatible with the conservative worst-case bound of \cref{eq:cost}, which already contains terms up to $O(p^4)$ before the dependence of $C_{\mathrm{fit}}(n,p)$ on model dimension is specified. It should therefore be interpreted only as an empirical scaling law over the range of expressions considered here, rather than as an asymptotic complexity result.

The main practical limitation is consequently scalability with model dimension, rather than a large constant-factor overhead specific to \texttt{FGP}. Runtime is dominated by the repeated nonlinear least-squares fits required for candidate evaluation and cross-validation, operations shared by all variants, which also explains their relatively similar wall-clock costs despite substantial differences in pruning behaviour. In this study we deliberately used iterative warm starts and conservative constant optimisation to reduce nonlinear fitting failures and isolate the behaviour of the pruning rules. Lighter optimisation schemes could substantially reduce runtime when occasional fitting failures can be tolerated.

\end{document}